\documentclass[runningheads,a4paper]{llncs}

\usepackage{amssymb}
\usepackage{graphicx}
\usepackage{subfigure}
\usepackage{caption}
\usepackage{url}
\usepackage[ruled,vlined]{algorithm2e}
\urldef{\mailsa}\path|{rhu8, Stephen.Watt}@uwo.ca|    
\newcommand{\keywords}[1]{\par\addvspace\baselineskip
\noindent\keywordname\enspace\ignorespaces#1}

\newcommand{\Paragraph}[1]{\vspace{-.5\baselineskip}\paragraph{#1}}

\begin{document}

\mainmatter  

\title{Determining Points on Handwritten Mathematical Symbols}

\titlerunning{Determining Points on Handwritten Mathematical Symbols}

\author{Rui Hu \and Stephen M. Watt}

\authorrunning{Determining Points on Handwritten Mathematical Symbols}

\institute{The University of Western Ontario\\
London Ontario, Canada N6A 5B7\\
\mailsa\\
}

\toctitle{Lecture Notes in Computer Science}
\tocauthor{Rui Hu}
\tocauthor{Stephen M. Watt}
\maketitle

\begin{abstract} 
In a variety of applications, such as handwritten mathematics and
diagram labelling, it is common to have symbols of many different
sizes in use and for the writing not to follow simple baselines. 
In order to understand the scale and relative positioning of
individual characters, it is necessary to identify the location of
certain expected features. These are typically identified by
particular points in the symbols, for example, the baseline of a lower
case ``p" would be identified by the lowest part of the bowl, ignoring
the descender. We investigate how to find these special points 
automatically so they may be used in a number of problems, such as
improving two-dimensional mathematical recognition and in handwriting neatening, while
preserving the original style.

\keywords{Handwriting analysis, Handwriting neatening, Mathematical handwriting recognition, Pen computing}
\end{abstract}

\section{Introduction}
Many digital ink applications allow 
handwritten characters in various sizes and in different locations. 
For example, in mathematics, subscripts and superscripts appear 
relatively smaller than normal text and are written slightly below 
or above it.  Moreover, these subscripts and superscripts may themselves have subscripts or superscripts.
Such notation is easily read and understood. This involves determining the relative baselines and sizes
of symbols.  This process may present various ambiguities, for example whether a particular symbol is
a lower case ``p'' or an upper case ``P'' giving the subscripted $p_q$ or the juxtaposed $Pq$.

In order to find the scale and offset of 
individual characters, it is necessary to identify the location of certain expected 
features which are typically defined by particular points. 
These particular points occur at different locations in different symbols, and the precise location can vary
in different handwriting samples of the same symbol. 
For example, the baseline of lowercase ``p" would be identified by 
the lowest part of the bowl, ignoring the descender. In contrast, the baseline of 
lowercase ``k", would be identified by the toes. In this article we refer to a point such as this, 
that determines the height of a metric line, as a \textit{determining point}. 
Knowing the determining points of each symbol can help us solve a number of problems. For example,
one can use the determining points to improve two-dimensional mathematical recognition. 
By comparing the baseline locations and the sizes of adjacent symbols, one can identify superscripts 
and subscripts (e.g. $S_2$, $S2$, $S^2$) with more confidence. Another application is in handwriting neatening. 
Since handwritten symbols often come with variations in alignment and size,
certain transformations based on determining points can be applied to obtain 
normalized samples while preserving the original writing style. 

Recording determining points for an individual handwritten symbol is easy. One can manually annotate the symbol
with the positions of all its determining points. However, finding determining points for all symbols in a collection
is much more challenging.   First, with a large database the labour for manual annotation would be prohibitively costly.
Secondly, applications such as mathematics 
involve a large variety of symbols derived from a range of alphabets and other sources.
In practice, many of them are often poorly written and there is no fixed dictionary 
of words to aid in disambiguation \cite{context-pen-based-computing}. 
This increases the difficulty to find determining points reliably. 
Meanwhile, each person's handwriting is unique ---
even identical twins write differently \cite{identical-twins-writing}. 
Even if a training database were to be fully annotated, it is not entirely clear how this 
should best be used to identify the points of interest
in new input.  
Last, but not least, the usual methods for detecting determining points 
depend on device resolution significantly. With rapidly evolving technology, this means that new algorithms cannot 
use archival data directly and therefore must be ``re-sampled'' (interpolated). 

We are interested in the problem of how to automatically find determining points
of handwritten mathematical symbols and to use them in a variety of problems. 
Considerable related work has been conducted, some of which we highlight here. 
Pechwitz and M\"argner \cite{skeleton} proposed an algorithm that can find determining 
points from symbol skeleton approximated by piecewise linear curve. However, these determining points 
are only useful in detecting baseline locations. 
In 2010, Infante Vel\'azquez \cite{TeresaThesis} developed an annotation 
tool to record determining points manually for handwritten 
characters represented in InkML \cite{inkml}. The determining points were later used to neaten new handwriting,
making it uniform in size, alignment and slant while preserving writers' particular writing styles. 
However, this tool recorded each determining point
with absolute coordinates and was therefore subject to device resolution and variations in style. 
As device resolution may vary among different vendors and over generations of technology, 
this approach is not device-independent. Similar problems exist in \cite{template-based-online-recognition}.
In addition, Zanibbi \textit{et al.} \cite{aidingmanipulation} proposed a technique to automatically improve the legibility of handwriting by gradually translating and scaling individual symbol to closely approximate their relative positions and sizes in a corresponding typeset version. This technique detects baseline locations by comparing symbols' bounding boxes, which leads to troubles with vertical placement and scale. For example, it fails to distinguish between ``$x2$" and ``$x^2$". 
In 2012, Hu and Watt \cite{Point-Selection} presented an algorithm to find turning 
points that determine the shape of characters, but that approach lacked the 
ability to capture the geometric meaning of each determining point and therefore does not 
provide sufficient information to calculate certain desired symbol metrics, such as the location of baseline. Harouni \textit{et al.} \cite{Deductive-Method-Finding-CP} later proposed a method to find determining 
points in handwritten Arabic characters. The method consisted of two stages. 
In the first stage, the raw input data were converted to a standard format using smoothing, 
normalization and interpolation techniques. 
In the second stage, each stroke of input characters was split into 
several pieces. The method calculated the local maximum and minimum of each piece and 
recorded them as determining points. However, this method is not optimal as it requires 
extra effort to split strokes and may generate undesired determining points that lack 
 meaning.

\begin{figure}[t]
\centering
\includegraphics[width=10cm]{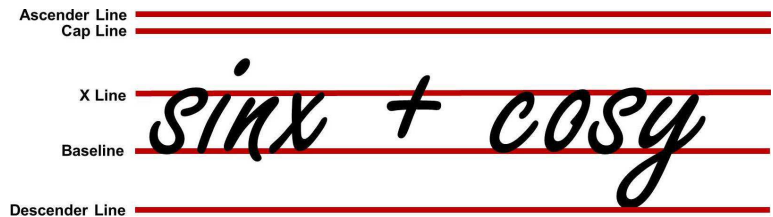}
\caption{An example to illustrate the concepts of metric lines.}
\vspace{-1.5\baselineskip}
\label{fig:metrics-lines}
\end{figure}

In this article, we present an algorithm to find determining points automatically and 
suggest how they may be used in areas such as improving two-dimensional 
mathematical recognition and in neatening handwriting.  The basic approach is to identify the points of 
interest on one average instance of each type of symbol, and to use this information to find the corresponding
points on newly written symbols. 
We borrow ideas from typography, where a number of determining 
points are identified to measure the metrics of different font families, and apply
these to handwriting.  We consider several types of determining points, which, in 
turn, determine certain metrics. These include the locations of the five main metric lines, i.e. the baseline, 
x line, ascender line, cap line, and descender line, as shown in Figure~\ref{fig:metrics-lines},
as well as symbol width and slant. To make the determining points device-independent, 
the algorithm first converts all handwritten symbols into parametric curves approximated by
truncated orthogonal series, mapping each symbol to a single point in a low dimensional vector space of series coefficients. 
We then compute the average symbol for each class by computing the average of the points for the class in the vector space.
The determining points of interest are identified on these average symbols. From these, 
the algorithm can derive corresponding determining points in samples automatically.
The beauty of this algorithm is that it is writer-independent. We only need to annotate once, 
on the average symbols. This reduces cost significantly. Furthermore, the algorithm is device-independent 
as all symbols are represented in the functional space, which is robust against changes in device
resolution.

The remainder of this article is organized as follows. In Section~\ref{sec:preliminaries},
we recall how to represent digital ink using functional approximation. 
Section~\ref{sec:metrics} discusses several types of determining 
points that are useful in finding symbol alignment lines. In Section~\ref{sec:algorithms}, we present the 
algorithm that can identify determining points in handwritten mathematical symbols automatically. 
Section~\ref{sec:experiments} evaluates the performance of the algorithm. 
We then investigate the possible use of the algorithm in a number of problems in Section~\ref{sec:use-cases}. 
Section~\ref{sec:conclusion} concludes the article.

\section{Functional Approximation for Digital Ink}\label{sec:preliminaries}
Digital ink is generated by sampling points from a traced curve at a certain rate, 
and thus is typically given in the form of a series of points, each of which 
contains {\it x} and {\it y} values in a rectangular coordinate system at a sequence of times. 
Since the sampling rate and resolution typically depend on the hardware type, different devices usually result in
different numerical point values for the same character. In order to take device differences into account,
various {\it ad hoc} treatments have been developed, such as size normalization and ``re-sampling'' (interpolation).
To make the representation more robust under changes in hardware, 
we represent handwritten symbols as coefficients for an approximating basis in a function space. 
This approach has been used in earlier work 
\cite{Polynomial-Approximation,Online-stroke-modeling,Succinct-Functional-Approximation,distance-based-classification}.

We consider an ink trace as a segment of a plane curve ($x(s)$, $y(s)$), 
parameterized by Euclidean arc length 
\vfill
$$s = \int \sqrt{dx^2+dy^2}.$$ 
\vfill
\noindent This parameterization has been found to lead to good recognition
and is intuitive sense, since it gives curves that look the same the
same parameterization \cite{Polynomial-Approximation}.
Given a digital ink trace $(x(s), y(s))$ and an approximating basis $\{B_i(s)\}_{i=0,\ldots,d}$, we represent the
trace using the coefficients $x_i$ and $y_i$ from 
\vfill
$$x(s) \approx \sum_{i=0}^d x_iB_i(s) \hspace{25pt} y(s) \approx \sum_{i=0}^d y_iB_i(s)$$
\vfill
\noindent It is convenient to choose the functions $B_i(s)$ to be orthogonal polynomials, e.g. 
Chebyshev, Legendre or some other polynomials. By choosing 
an appropriate family of basis polynomials to high enough degree, 
the approximating curve can be made arbitrarily close to the original trace.

\begin{figure}[t]
\centering
\subfigure[]{
\includegraphics[width=3.8cm]{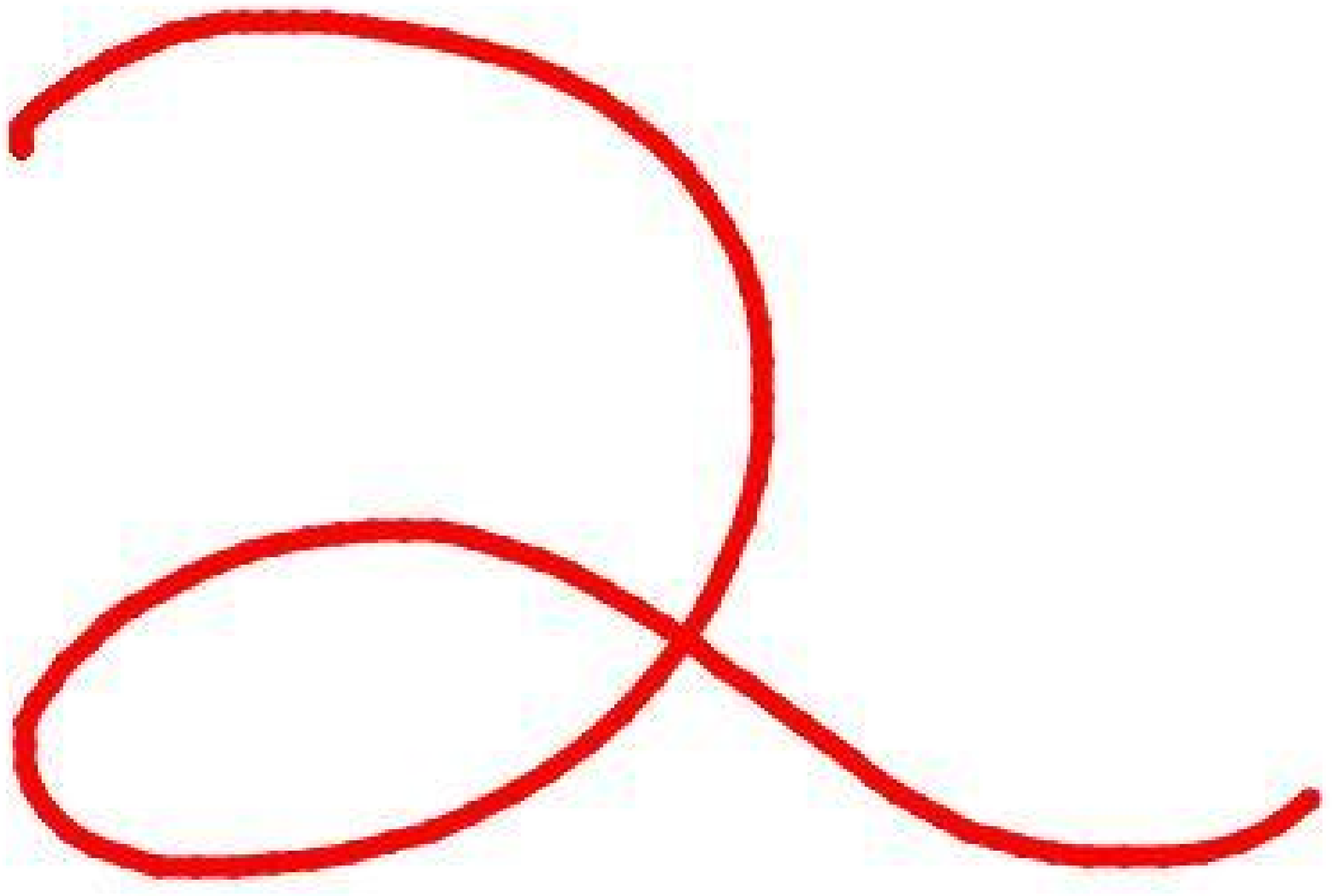}
\label{fig:original-character}
}\hspace{1.6mm}
\subfigure[]{
\includegraphics[width=3.8cm]{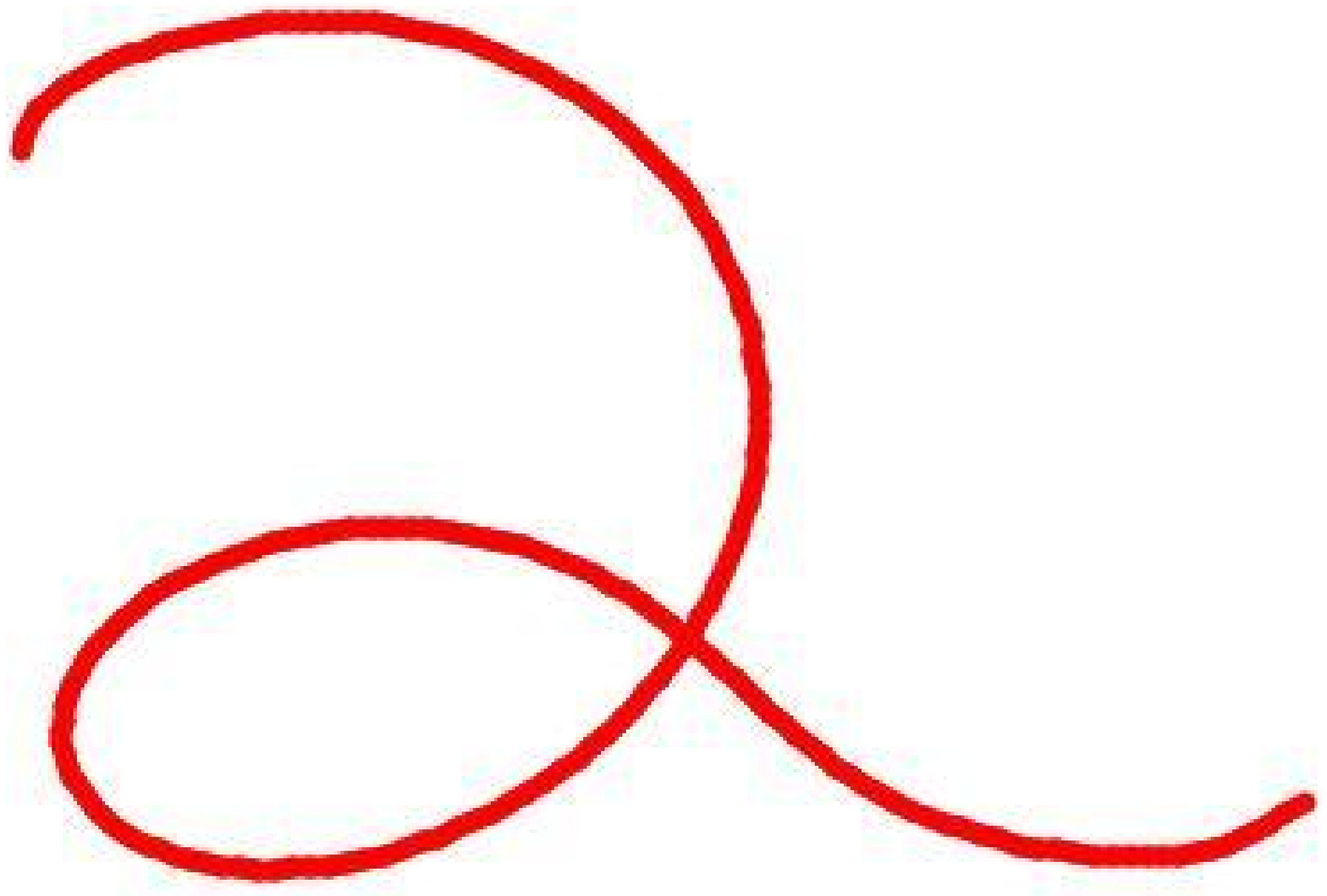}
\label{fig:approximated-character}
}
\caption[]{Approximation using Legendre-Sobolev series. (a) Original. \newline (b) Approximated using series of degree 12 with $\mu = 1/8$.}
\label{fig:lsp-approximation}
\end{figure}

We have found a Legendre-Sobolev basis allows approximating curves to have the desired shape for relatively low degrees.
These  may be computed by Gram-Schmidt orthogonalization of the monomials $\{s^i\}$ with respect to the inner product 
\vfill
$$\langle f, g \rangle = \int_a^b f(s) g(s) \mathrm{d}s + \mu \int_a^b f^\prime(s) g^\prime(s) \mathrm{d}s.$$
\vfill
If a symbol has multiple strokes, we join consecutive strokes by concatenating the point series, 
which yields a single curve.  For more details see~\cite{distance-based-classification}.
An example of using Legendre-Sobolev polynomials in approximation is
shown in Figure~\ref{fig:lsp-approximation}. 
After approximation, we may now represent the digital ink trace, or symbol, as 
the coefficient vector $(x_0, ..., x_d, y_0, ..., y_d)$. We may standardize the location and size
of the character by setting $x_0, y_0$ to 0 and the norm of the vector to 1.
\section{Handwriting Metrics}\label{sec:metrics}
In order to 
understand the scale of individual symbols, it is necessary to identify the location
of certain expected features which are typically defined by a number of determining points.
These determining points have locations that vary from symbol to symbol, but typically occur where parts of the
symbols touch certain invisible horizontal lines.
To discuss this, we adopt concepts from typeface design. 
In this article, we consider several types of determining points related to the following 
metrics.  We concentrate on symbols used in European alphabets.  Many other writing systems would have other metric lines determined in a similar way.

\begin{figure}[t]
\centering
\vspace{4mm}

\subfigure[]{
\includegraphics[width=4.8cm]{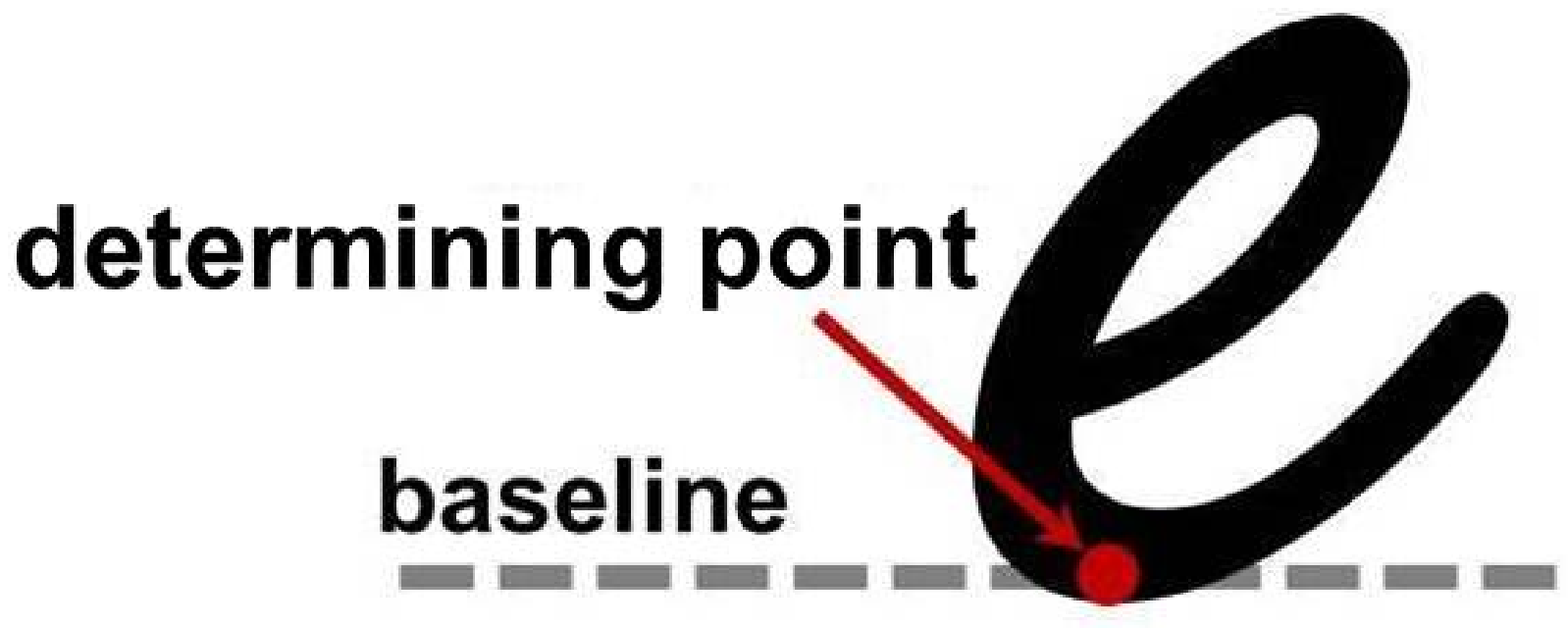}
\label{fig:baseline1}
}\hspace{1mm}
\subfigure[]{
\includegraphics[width=4.8cm]{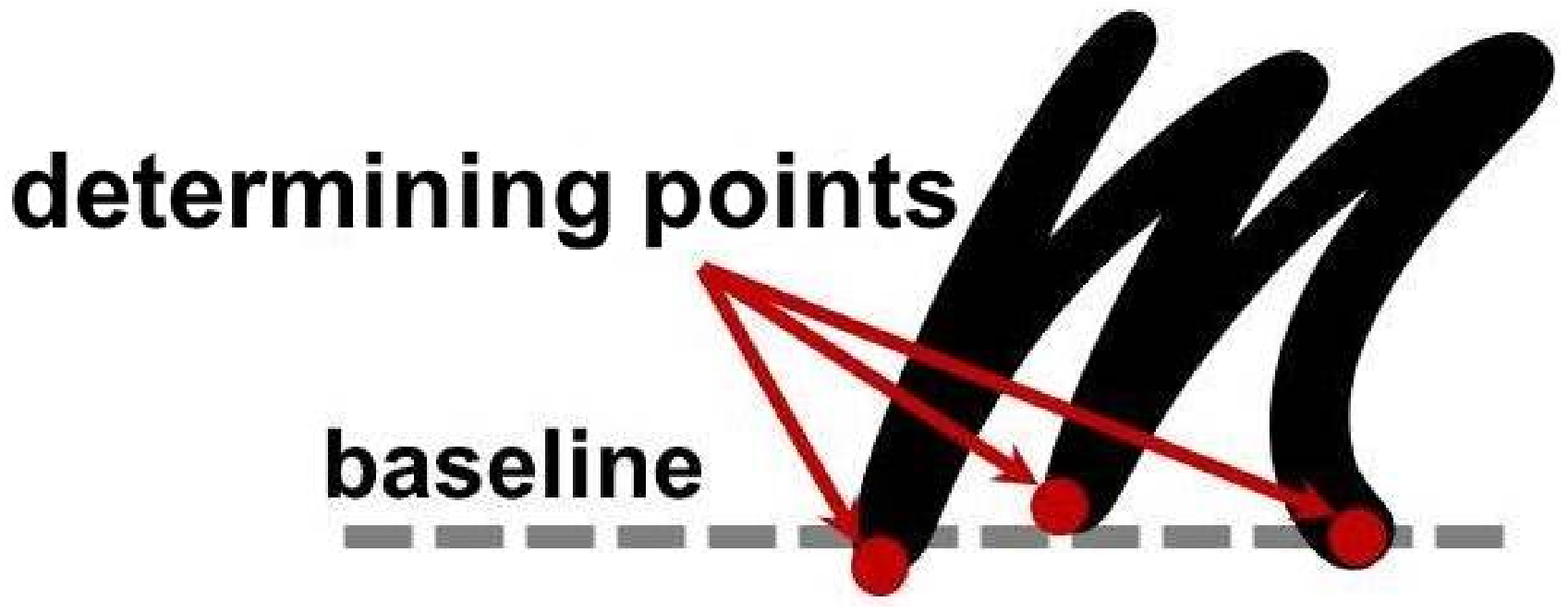}
\label{fig:baseline2}
}
\caption[]{Baseline with (a) one, and (b) multiple determining points.}
\label{fig:baselines}
\end{figure}

\Paragraph{\textbf{Baseline}}
Most scripts share the notion of {\it baseline}. It is a guide 
line for writing so that adjacent symbols can retain their horizontal alignment. It is also used as 
the reference to obtain other metrics such as x height, ascender height, etc.
While some symbols such as lower
case ``p" may extend below the baseline, it serves as the imaginary base for most symbols. 
Figure~\ref{fig:baselines} shows examples of baselines and their
determining points. As shown in Figure~\ref{fig:baseline2}, the three legs of the lowercase ``m" are not 
completely aligned. In such case, multiple determining points are identified and the location 
of the baseline may be determined by the average $y$ value of all the determining points.

\begin{figure}[t]
\centering
\subfigure[]{
\includegraphics[width=4.5cm]{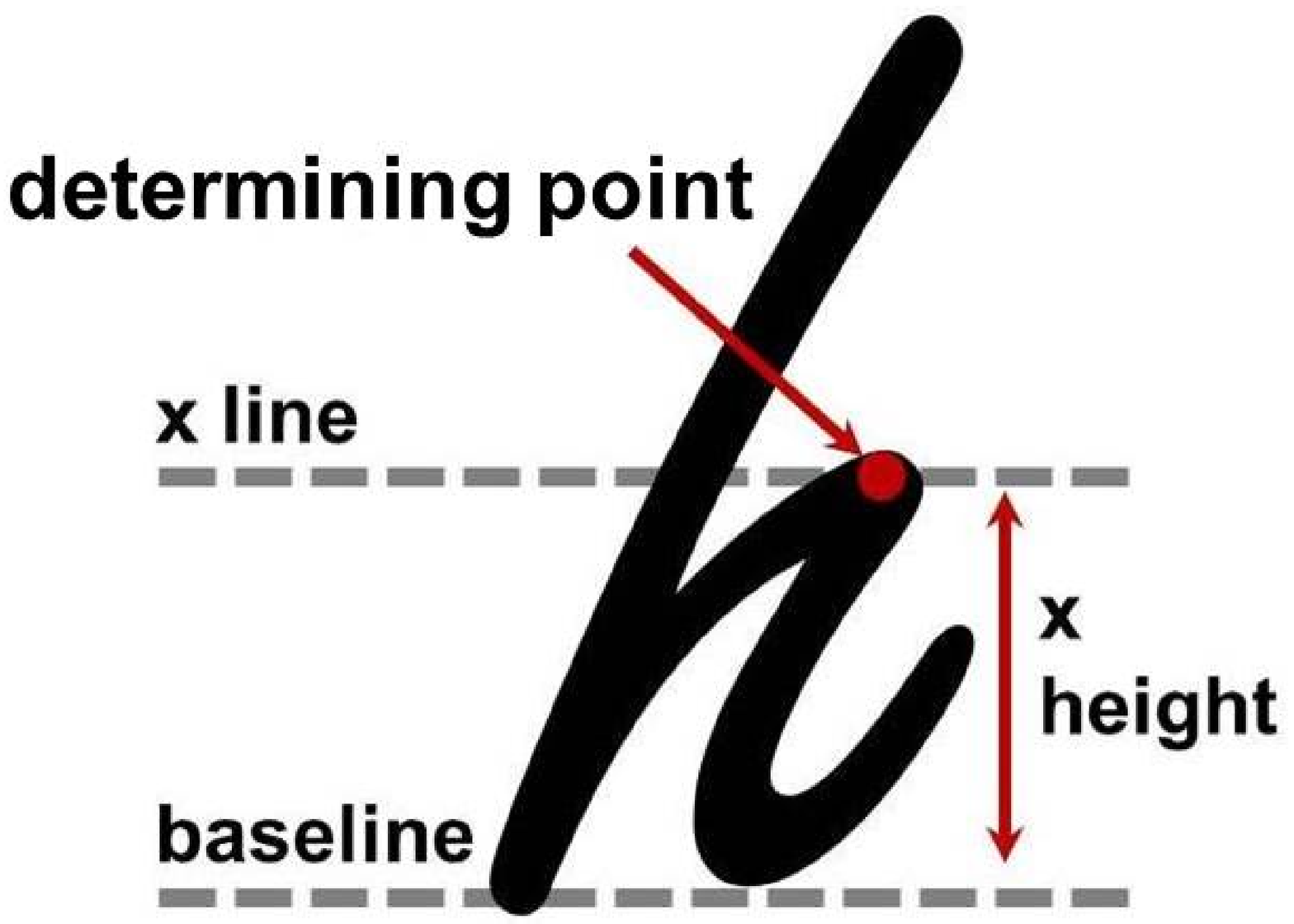}
\label{fig:x-line-height}
}\hspace{1mm}
\subfigure[]{
\includegraphics[width=4.5cm]{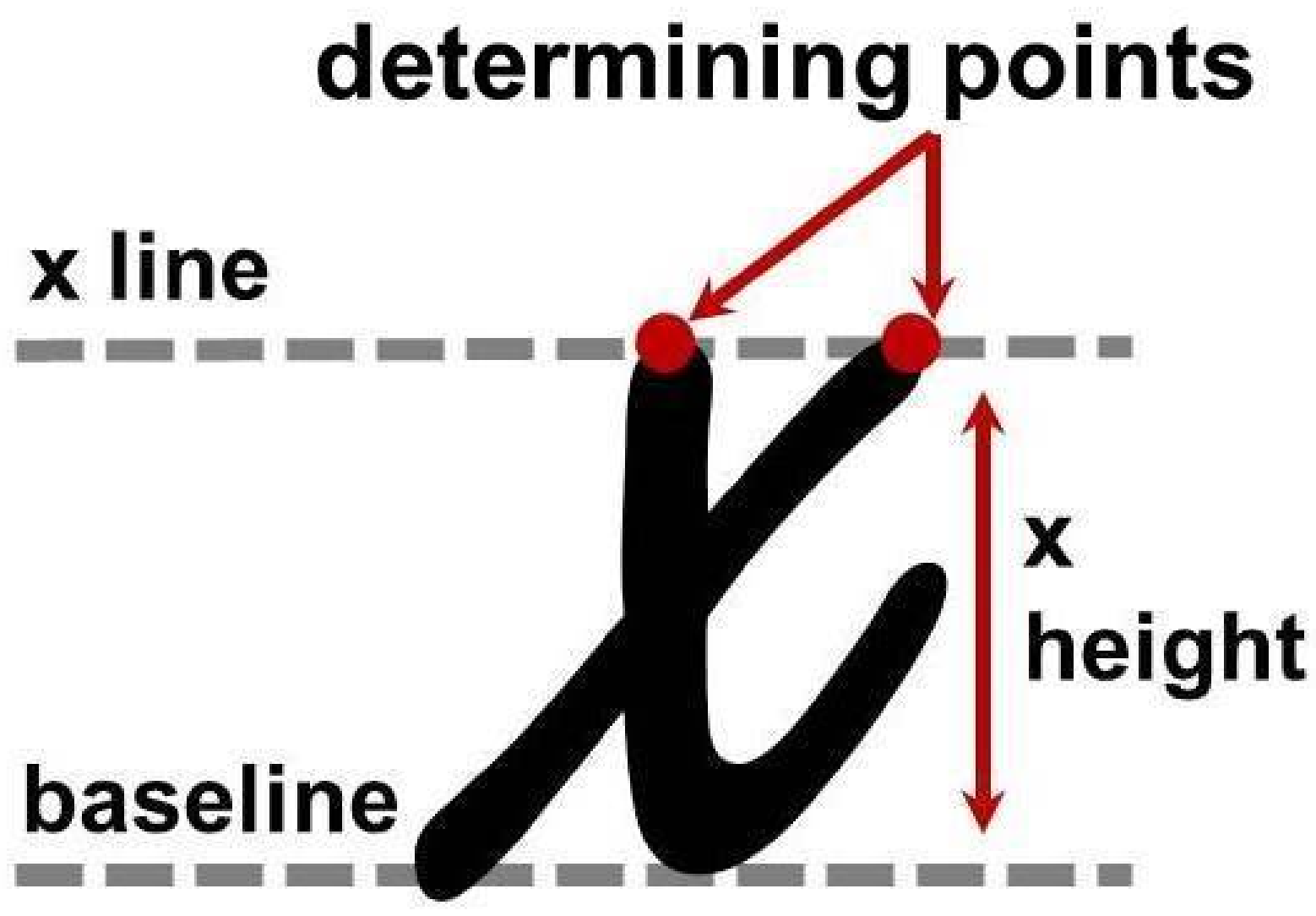}
\label{fig:x-line-height2}
}
\caption[]{x line and x height with (a) one, and (b) multiple determining points.}
\label{fig:x-line-heights}
\end{figure}

\Paragraph{\textbf{X Line and Height}}
The {\it x line} falls at the top of most lowercase symbols, such as ``a" and ``y", 
and is located over the baseline. Some symbols may extend above the x line, such as ``h" where
the  x line is located at the top of the shoulder. The {\it x height} is the distance between 
the {\it baseline} and the {\it x line}. Figure~\ref{fig:x-line-heights} shows an 
example of x line and associated determining points. 
Certain symbols, such as lowercase ``x", may have multiple determining points to 
define the x line. In such a case, the location of the x line 
is determined by the average of their $y$ values.

\begin{figure}[t]
	\begin{minipage}[t]{0.5\linewidth}
	\centering
	\includegraphics[width=4.5cm]{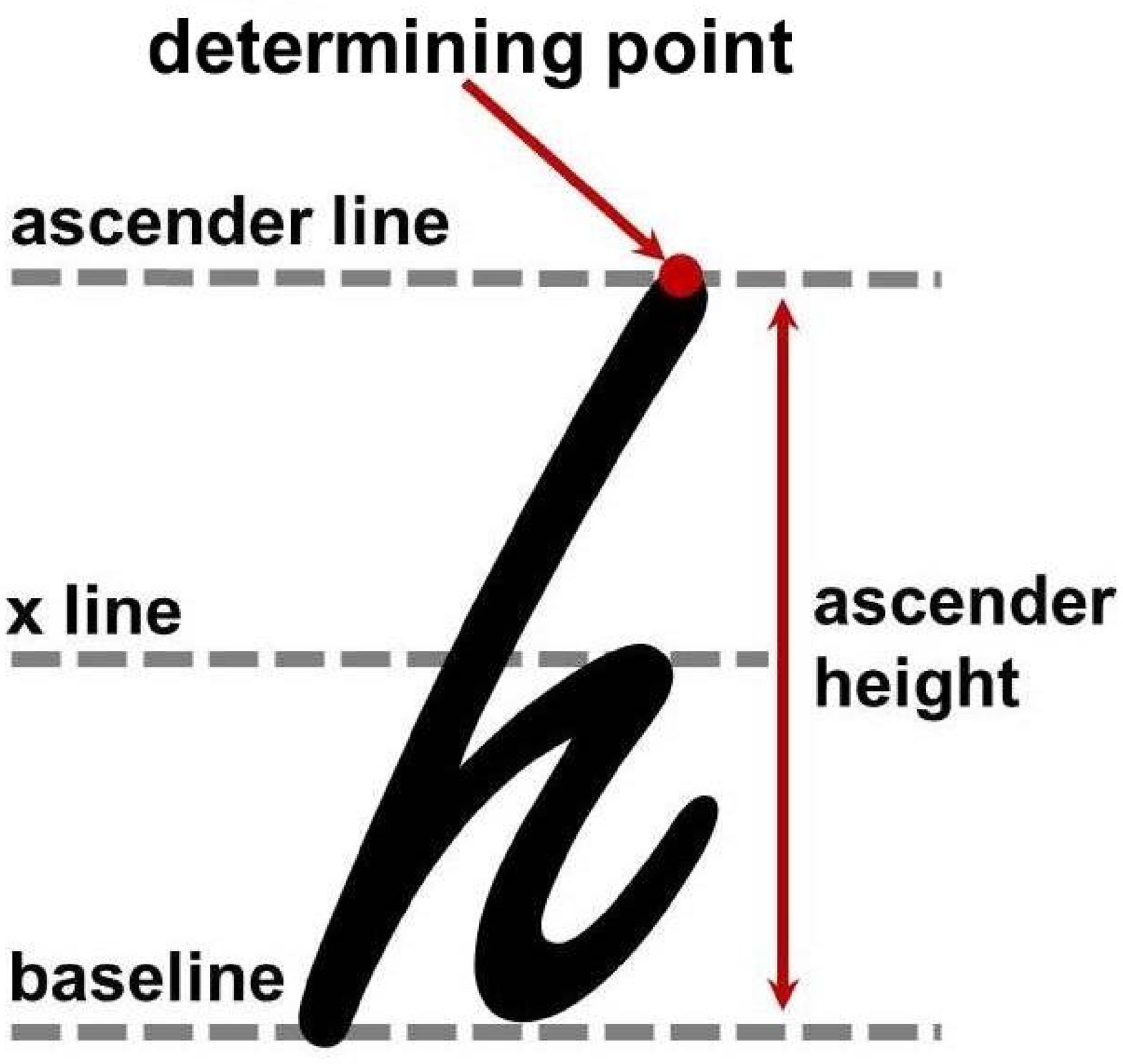}
	\caption{Ascender line and height}
	\label{fig:ascender-line}
	\end{minipage}
	\hspace{0.1cm}
	\begin{minipage}[t]{0.45\linewidth}
	\centering
	\includegraphics[width=4.3cm]{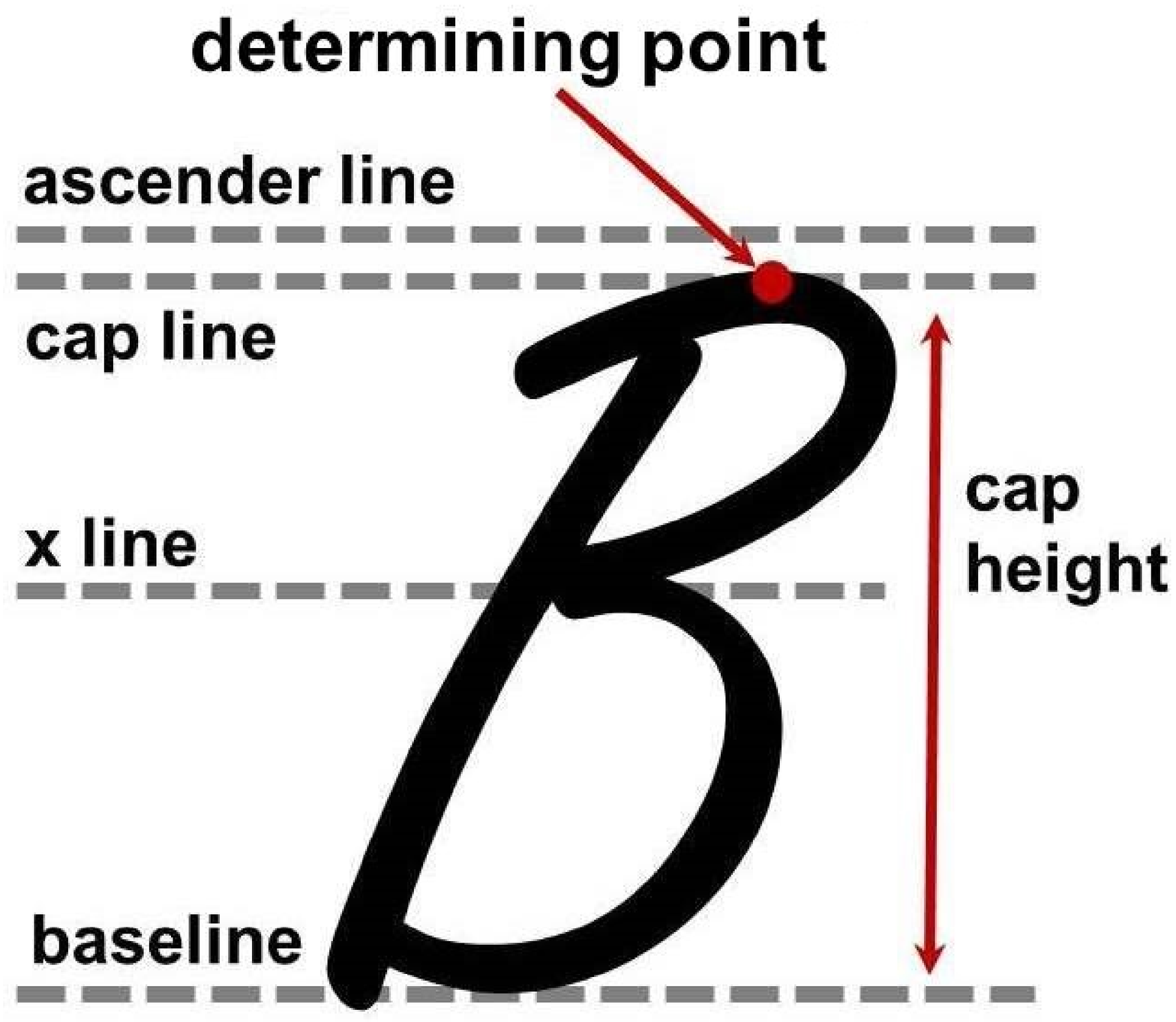}
	\caption{Cap line and height}
	\label{fig:cap-line}
	\end{minipage}
\vspace{\baselineskip}
\end{figure}

\Paragraph{\textbf{Ascender Line and Height}}
The part of a lowercase symbol, such as ``h" and ``k", that extends above the {\it x line} is 
known as an {\it ascender}. The {\it ascender line} is located above the {\it x line} and is 
determined by the height of the ascenders. The {\it ascender height} is the distance between the
baseline and the ascender line. Figure~\ref{fig:ascender-line} shows an example of an ascender line and ascender height.
The location of the ascender line is determined by the determining point shown in red. In the case that 
there are multiple determining points, the location of the ascender line is given by the
average $y$ value of all the relevant determining points.

\Paragraph{\textbf{Cap Line and Height}}
The {\it cap line} is used to align uppercase symbols and is usually located below the ascender line, although it is not limited to that position.  
Indeed, in handwriting it often coincides with the ascender line.
The {\it cap height} is the distance between the baseline
and the cap line. Figure~\ref{fig:cap-line} shows an example of a cap line and cap height.
The location of the cap line is determined by the determining point shown in red. 
In the case that there are multiple determining points, the location of the 
cap line may be taken as the average $y$ value of all the determining points.

\begin{figure}[t]

	\begin{minipage}[t]{0.46\linewidth}
	\centering
	\includegraphics[width=4.8cm]{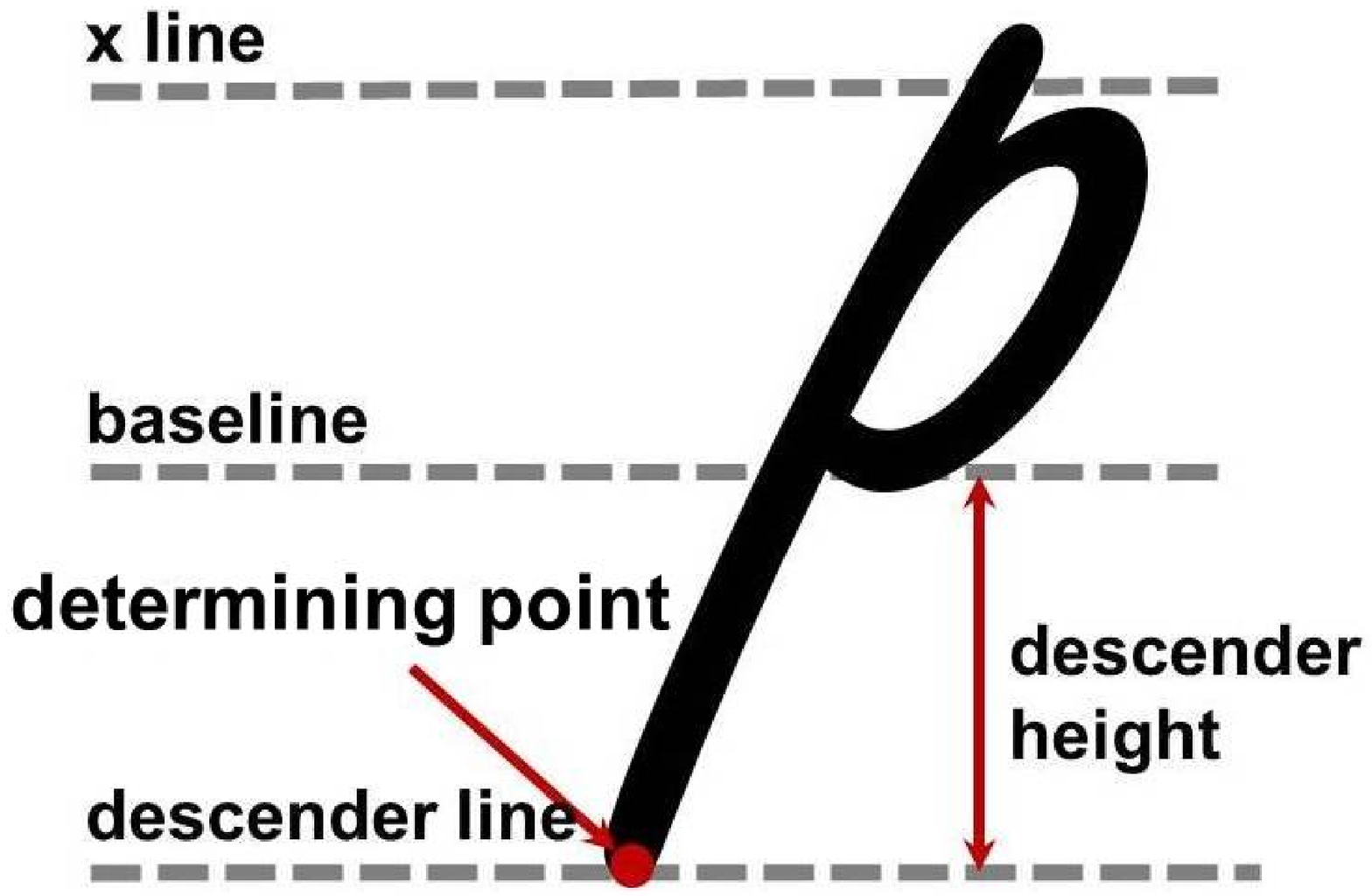}
	\caption{Descender line and height}
	\label{fig:descender-line}
	\end{minipage}
	\hspace{0.1cm}
	\begin{minipage}[t]{0.44\linewidth}
	\centering
	\includegraphics[width=3.6cm]{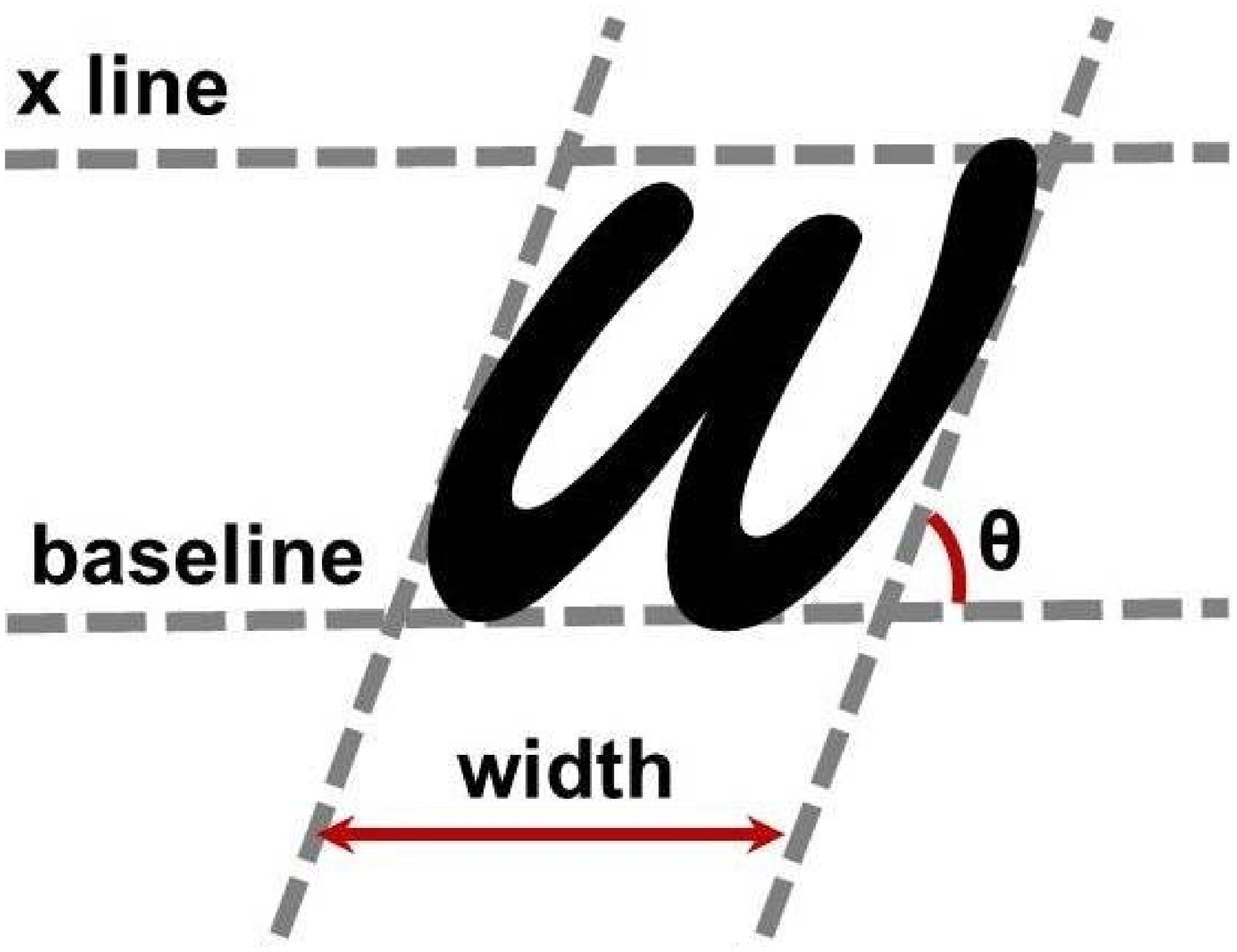}
	\caption{Slant ($\theta$) and width}
	\label{fig:slant-width}
	\end{minipage}
\vspace{\baselineskip}
\end{figure}

\Paragraph{\textbf{Descender Line and Height}}
The {\it descender line} is located below the baseline. It is used to align
 descenders, which are the parts of symbols that extend below the baseline.
Figure~\ref{fig:descender-line} shows an example of a descender line and descender height.
If there are multiple determining points, the location of the 
descender line is given by the average $y$ value of all the determining points.

\Paragraph{\textbf{Slant and Width}}
In some handwriting styles, symbols are written with inclination either to the left
or to the right. The degree of inclination is referred to as the {\it slant}. The {\it width} of a symbol is given by
 the horizontal distance from the left-bounding and right-bounding lines with the given slant. 
Figure~\ref{fig:slant-width} shows an example of symbol width and slant.

\section{Algorithm}\label{sec:algorithms}

In this section, we present an algorithm to find automatically the determining points for newly written symbols. 
The algorithm derives determining points for a new symbol from the known determining points of an annotated average symbol of the same type.

\subsection*{Average Symbols}
We classify symbols so that symbols that are written the same way and could be interpreted the same way are in the same class.  So, for example,
there may be several classes for the numeral ``8'', depending on whether the symbol is written with one continuous stroke or two separate strokes, which stroke is written first and the direction of writing.   On the other hand, a Latin letter ``O'' and the numeral ``0'' could belong to the same class.   

\begin{figure}[t]
\centering
\subfigure[]{
\includegraphics[width=9cm]{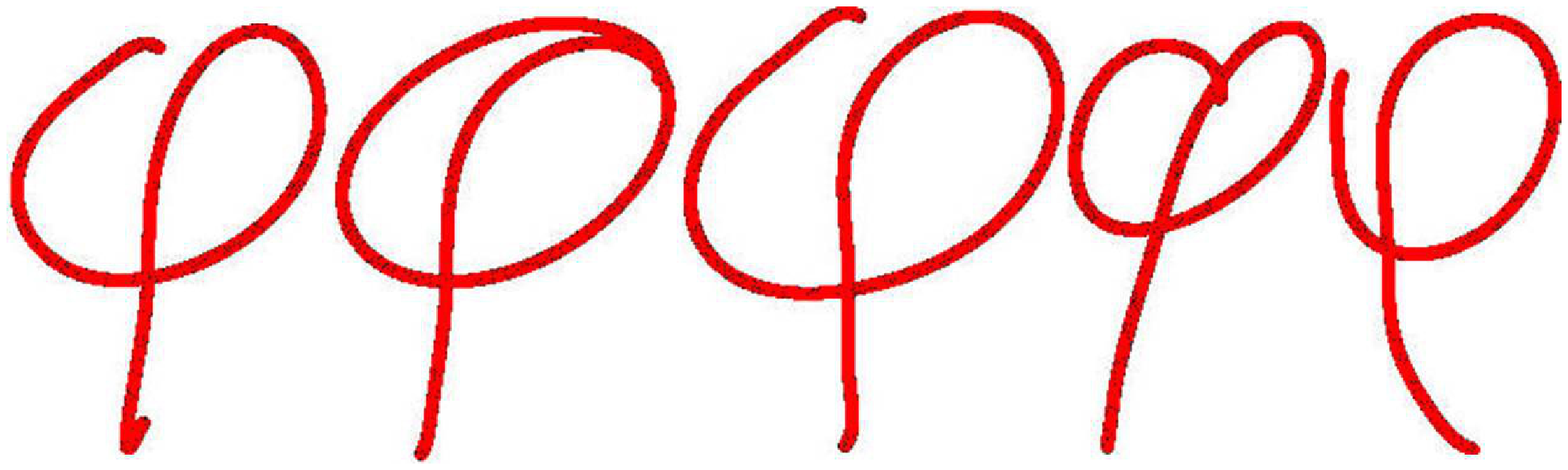}
\label{fig:Phi-Samples}
}\hspace{8mm}
\subfigure[]{
\includegraphics[width=1.72cm]{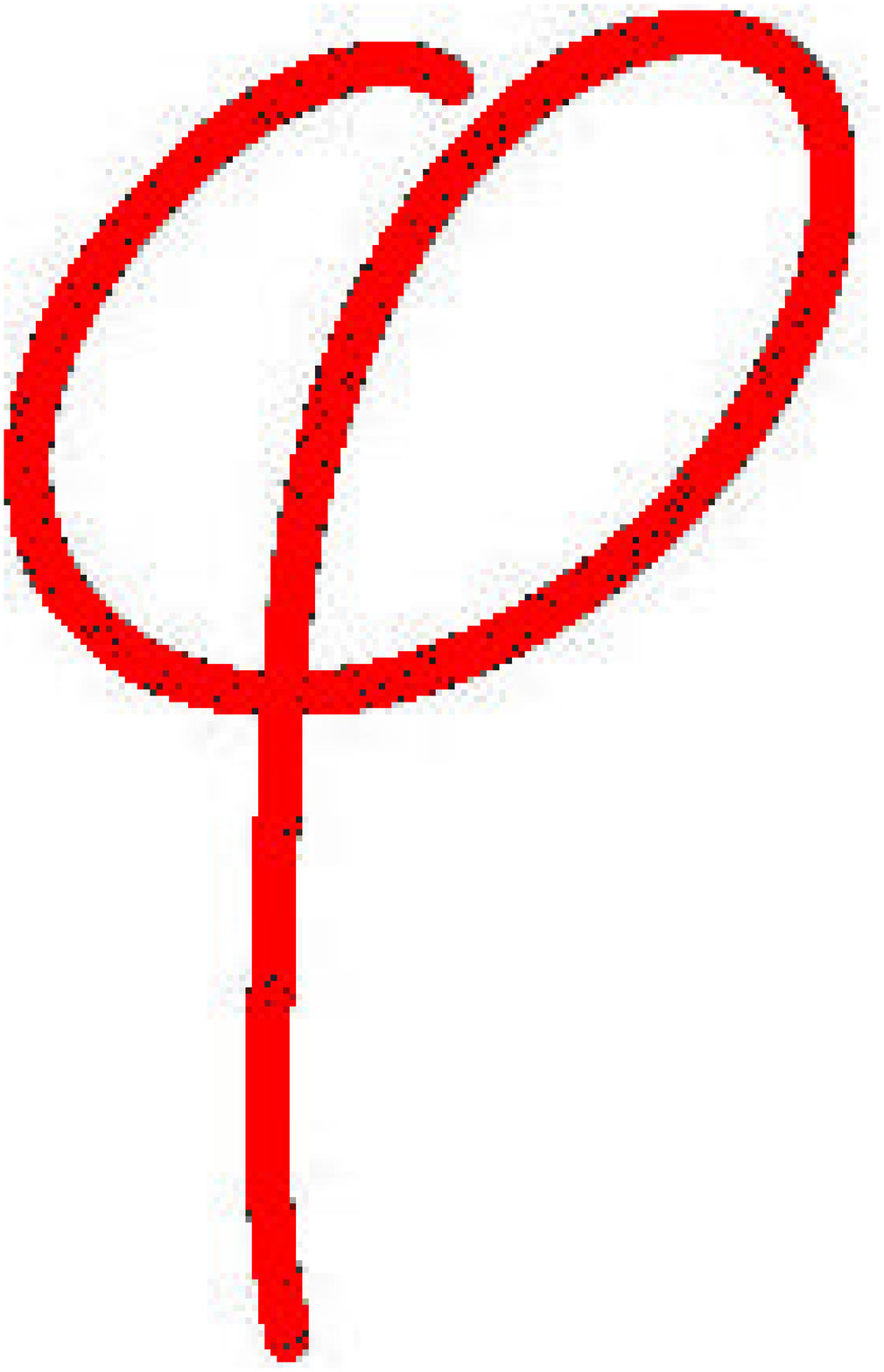}
\label{fig:Phi-Average}
}
\caption[]{(a) Samples provided by different writers. (b) The average symbol.}
\label{fig:average-symbol}

\end{figure}

Taking each sample as a point in the functional approximation space, it has been found in earlier work that classes of points are almost completely pair-wise separable by single hyperplanes.  Thus the convex hulls of the class point sets are to a good approximation non-intersecting. Any point on a line segment between two sample points of the same class falls within that class. It is therefore meaningful to compute the average of a set of known samples for a class as the average point in the function space
$$\bar C = \sum_{i=1}^n C_{i}/{n},$$
where $n$ is the number of the samples and $C_i$ is the coefficient vector for the $i^{th}$ sample. 
Figure~\ref{fig:Phi-Samples} shows a set of samples provided by different writers and Figure~\ref{fig:Phi-Average} shows
the average symbol.

\subsection*{Deriving Determining Points from Average Symbols}
Our algorithm is based on the observation that the average symbols typically
look similar to the samples of the same class. Within a given class, the features present in one sample should be present  in other samples and at a similar location.
We can take the location to be the arc length along the ink trace to the defining point of the feature.   We assume that, if two symbols are sufficiently similar, the locations of corresponding determining points will be similar (given by distance along the curve). 

This suggests that we can find the determining points of a new symbol by taking the known locations on an annotated symbol and making an adjustment.
In more detail, to detect the determining points in a sample, we start with an annotated sample in the same class.  For now, this will be the average of the training samples, annotated with its determining points.   Each annotation consists of the location (as arc length), the type of determining point (e.g. baseline, x line, etc) and whether it is located at a local minimum or local maximum of $y$ value.

For each determining point of the annotated sample, we guess that the corresponding determining point on the new sample will be near the same arc length location.   So we take the point at that location in the new sample and follow the trace upward or downward, depending on whether that determining point is supposed to be at a local
minimum or local maximum.
This can be easily done using a number of numerical methods. 
In our implementation, we applied Newton's method to solve $y'(s) = 0$. A formal 
algorithm is given in Algorithm~\ref{algo:LocateDeterminingPoints}.

\begin{algorithm}
\SetKwInOut{Input}{Input}
\SetKwInOut{Output}{Output}
\Input{$A$, the coefficient vector for a reference symbol.
\newline\newline
$D_A= [(s_1, T_1, K_1), \ldots (s_n, T_n, K_n)]$, 
a vector of determining points. \newline
For each, the position is given as arc length $s_i$ on the curve of $A$,
the value $T_i$ states which type of metric line is being defined, and
the value $K_i$ states whether the metric line is given by a local
minimum or local maximum at $y_A(s_i)$.
\newline\newline
$S$, the coefficient vector for the input sample whose determining points are to be found.}
\BlankLine
\Output{$D_S = [(\ell_1, T_1, K_1), \ldots, (\ell_n, T_n, K_n)]$, giving the locations, $\ell_i$, and types of the determining points of $S$.   \newline
The value of $\ell_i$ along $S$ corresponds to the value $s_i$ along $A$.}
\BlankLine

\BlankLine
1. Let $x_A(s)$, $y_A(s)$, $x_S(s)$, $y_S(s)$ be the coordinate functions defined
by the coefficient vectors $A$ and $S$.

2. \For{$i \in 1..n$}{

\If{$K_i = \textbf{max}$}{
    $f \longleftarrow -y_S$
}
\ElseIf{$K_i = \textbf{min}$}{
    $f \longleftarrow y_S$
}
$\ell_i \longleftarrow$ local minimum of $f(s)$ nearest $s = s_i$.\newline
    
\textit{Note this local minimum or maximum is of a real univariate polynomial and any standard method may be used. For example, we use Newton's method to solve $f^\prime(s)=0$ with initial point $s = s_i$.}

}
3. Return$[(\ell_1, T_1, K_1),\ldots, (\ell_n, T_n, K_n)]$
\caption{LocateDeterminingPoints}
\label{algo:LocateDeterminingPoints}
\end{algorithm}

\begin{figure}[t]
\centering
\subfigure[]{
\includegraphics[width=2.5cm]{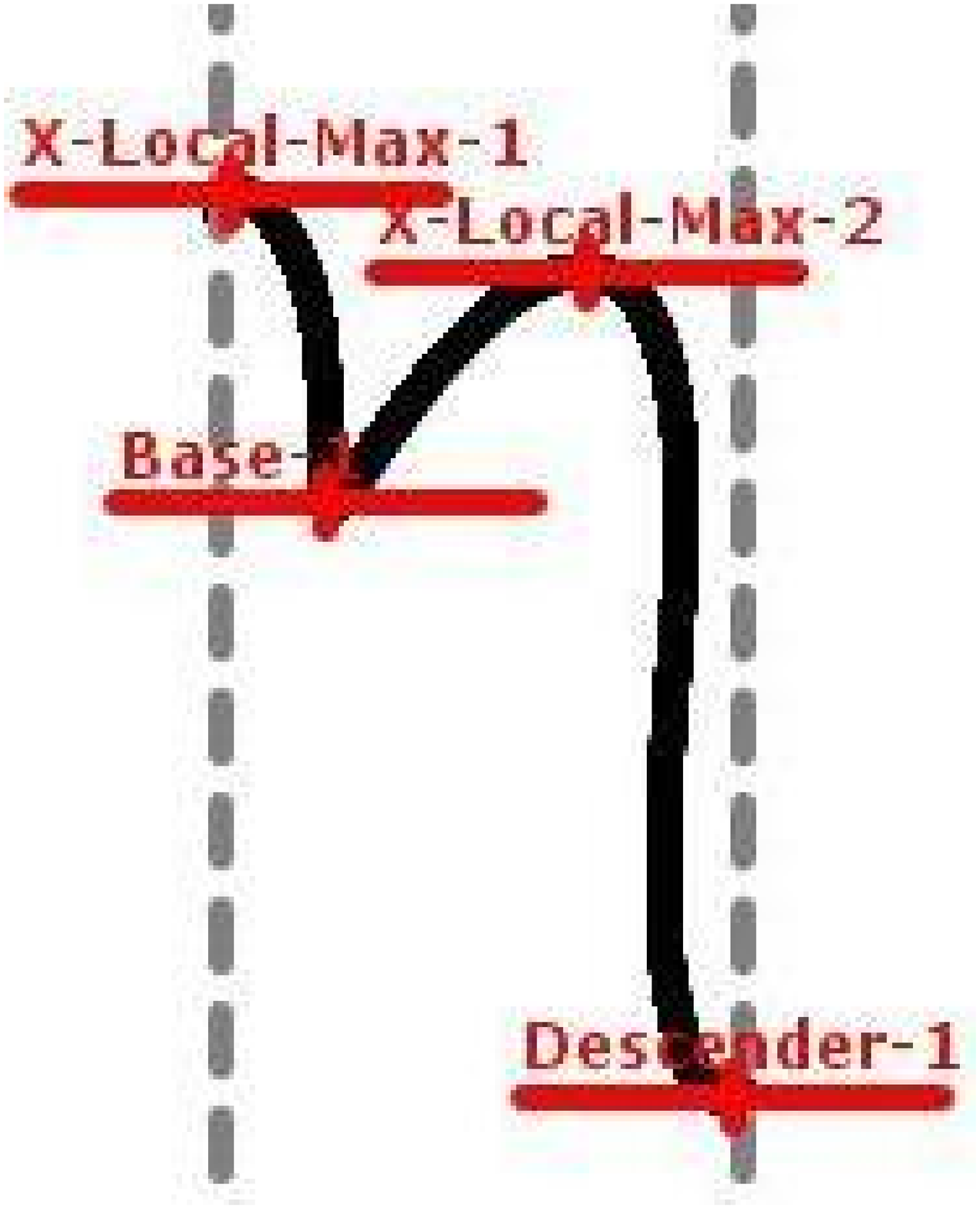}
\label{fig:Average-eta}
}\hspace{-3mm}\renewcommand{\thesubfigure}{(b1)}
\subfigure[]{
\includegraphics[width=2.3cm]{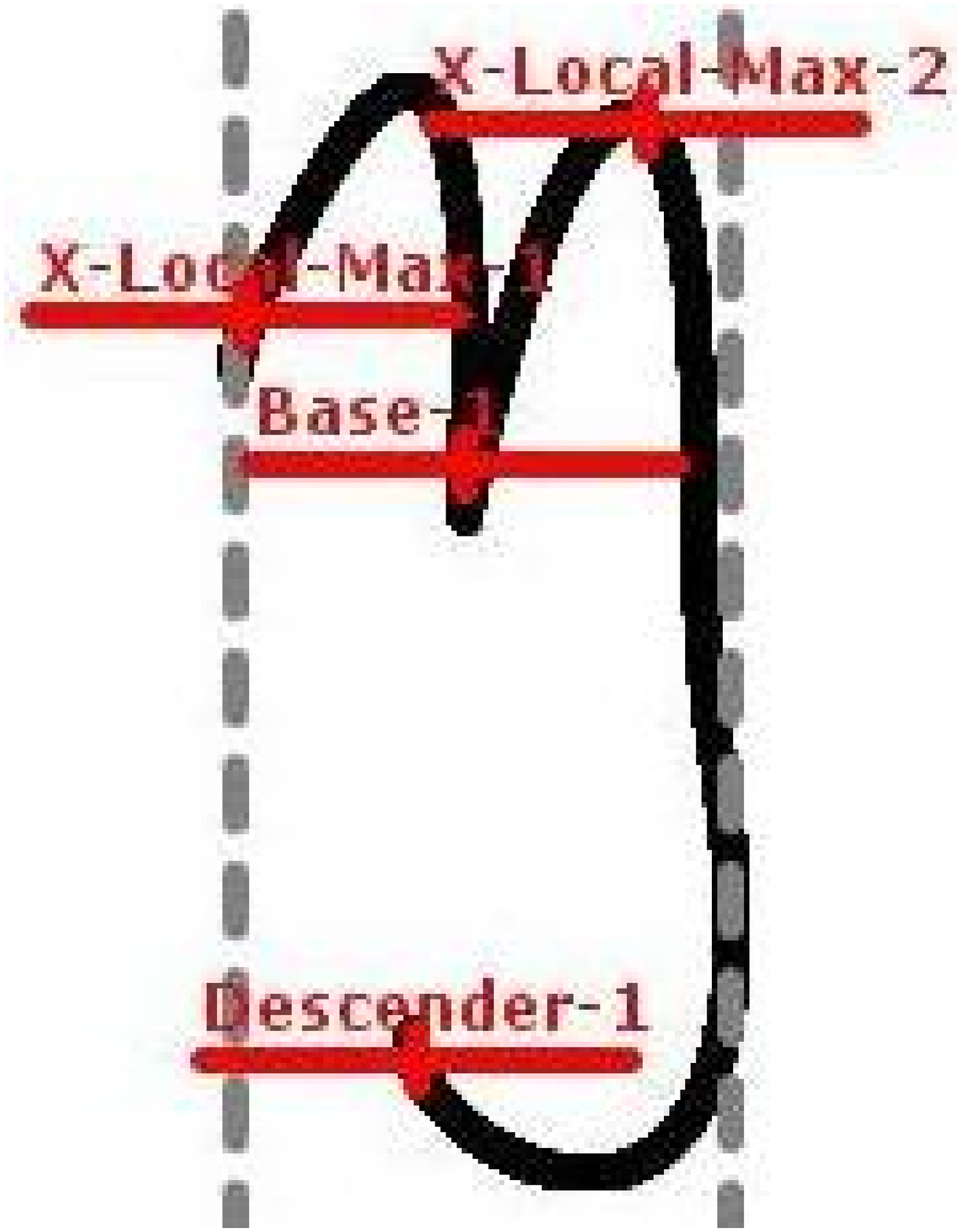}
\label{fig:eta-1-initial}
}\renewcommand{\thesubfigure}{(b2)}
\subfigure[]{
\includegraphics[width=1.9cm]{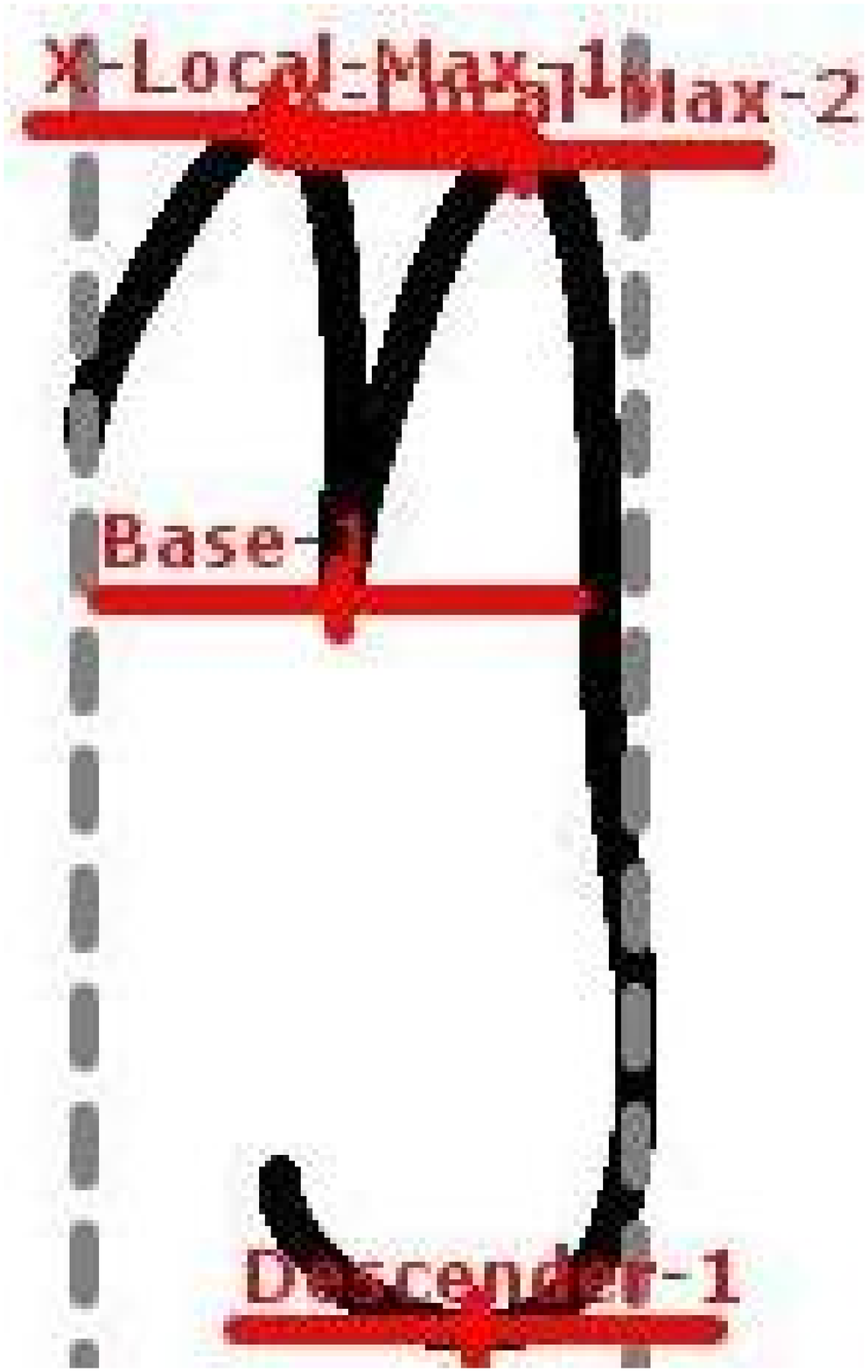}
\label{fig:eta-1-derived}
}\hspace{-2mm}\renewcommand{\thesubfigure}{(c1)}
\subfigure[]{
\includegraphics[width=2.15cm]{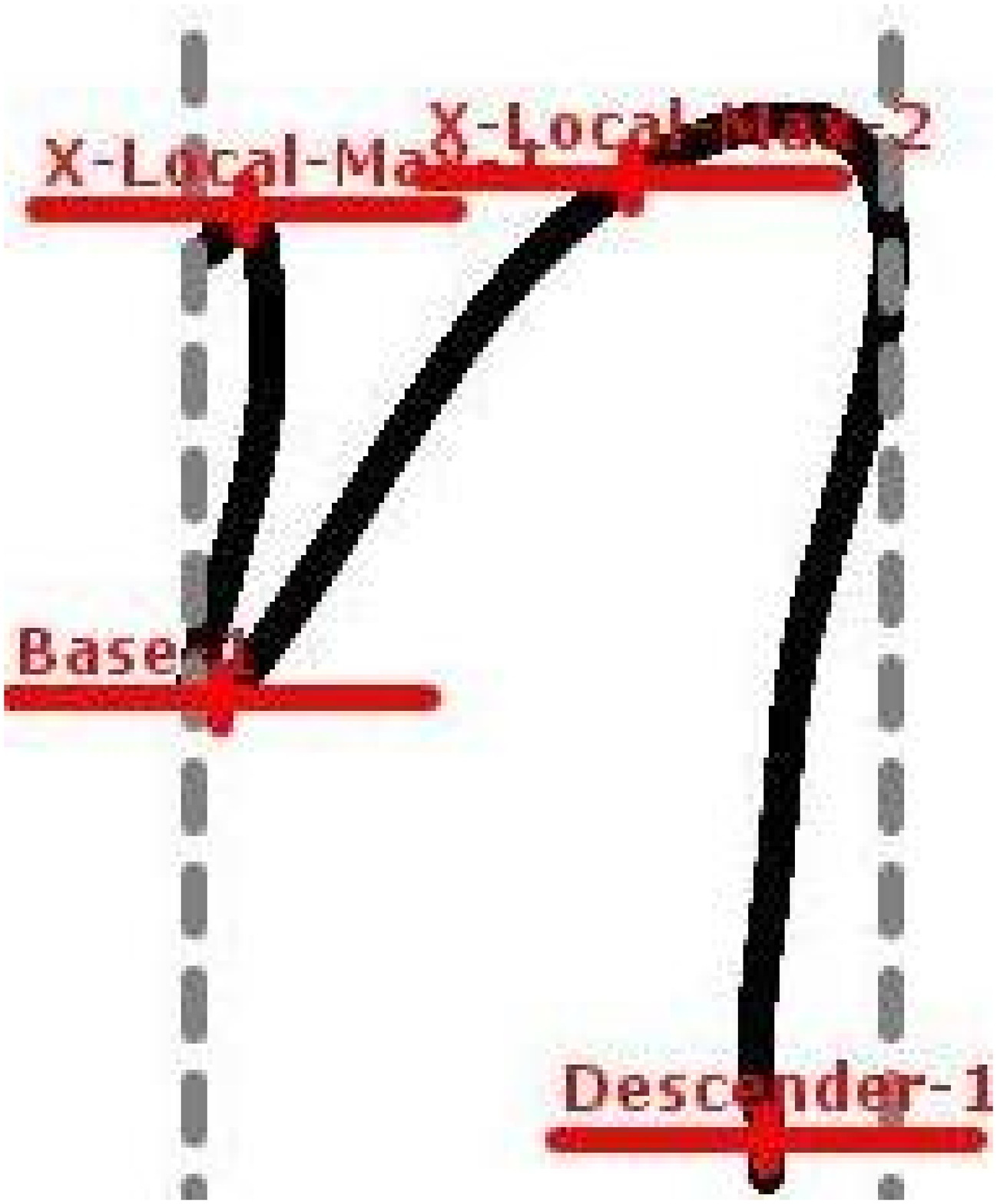}
\label{fig:eta-2-initial}
}\renewcommand{\thesubfigure}{(c2)}
\subfigure[]{
\includegraphics[width=2.3cm]{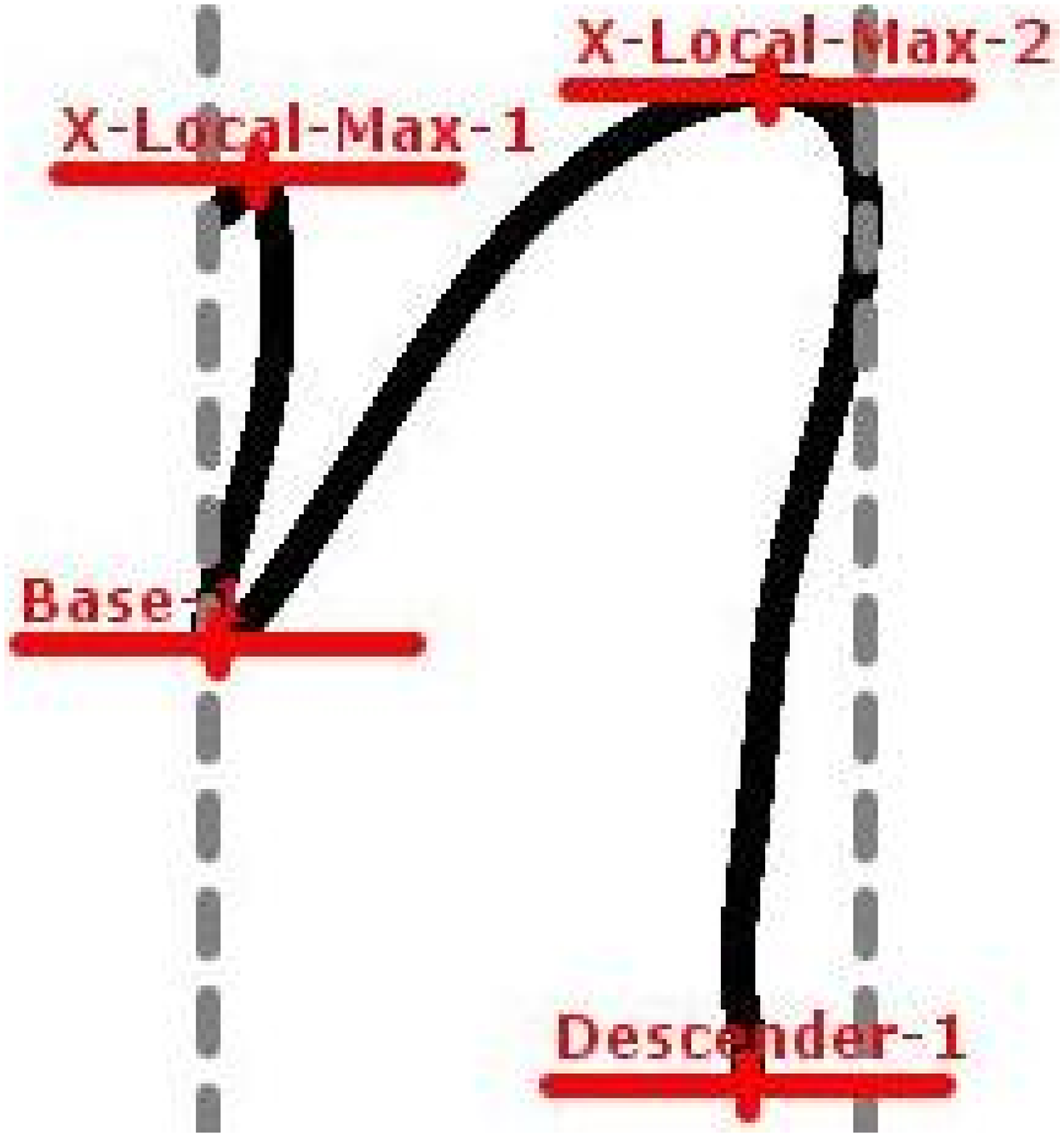}
\label{fig:eta-2-derived}
}
\caption[]{Automatically finding determining points. 
(a) Average symbol ``$\eta$". \newline
(b1) Sample 1 initial approximations and (b2) with determining points found. \newline
(c1) Sample 2 initial approximations and (c2) with determining points found.}
\label{fig:eta-determining-points}
\end{figure}

\begin{figure}[t]
\centering
\subfigure[]{
\includegraphics[width=2.1cm]{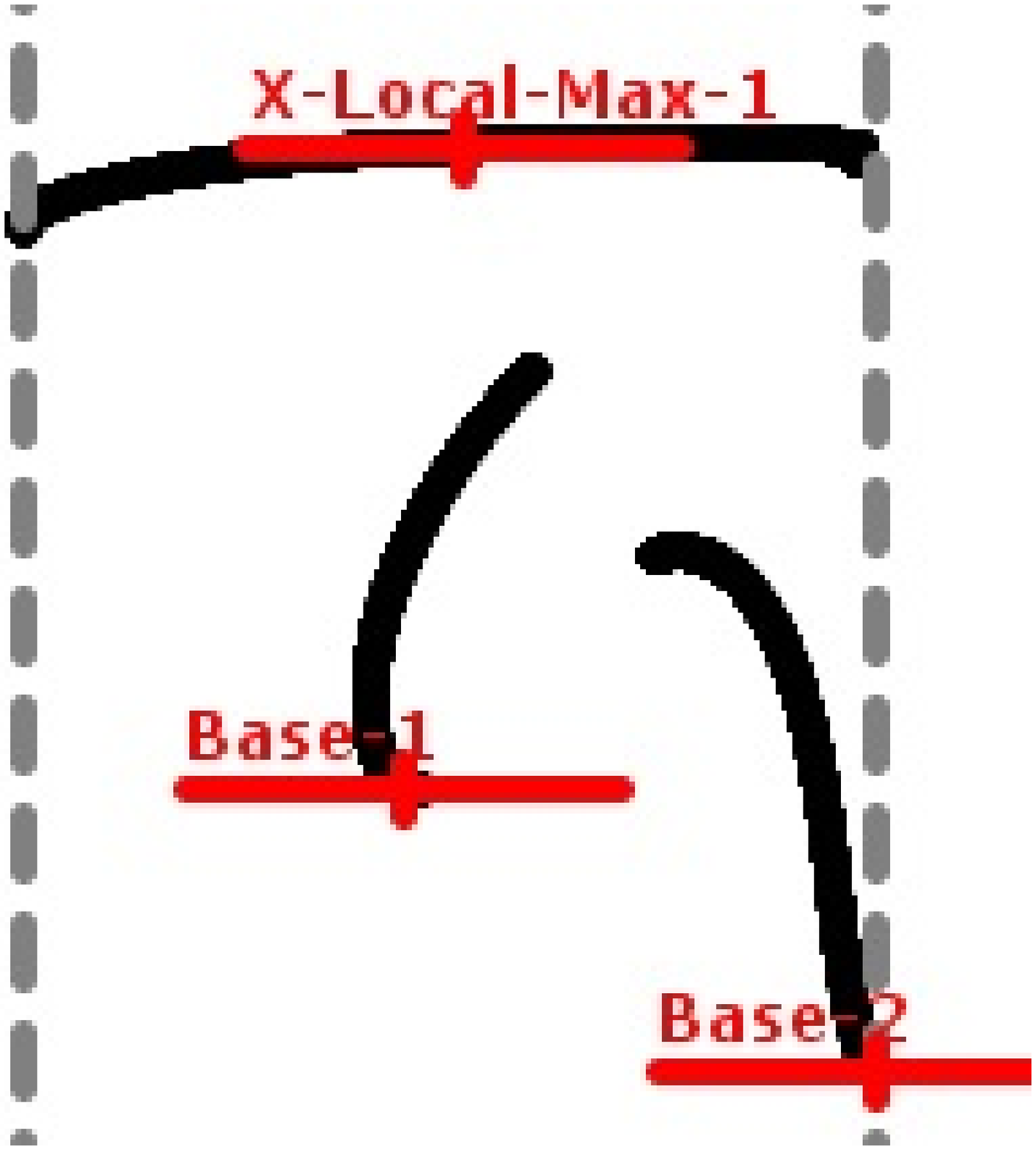}
\label{fig:Average-P}
}
\subfigure[]{
\includegraphics[width=2.1cm]{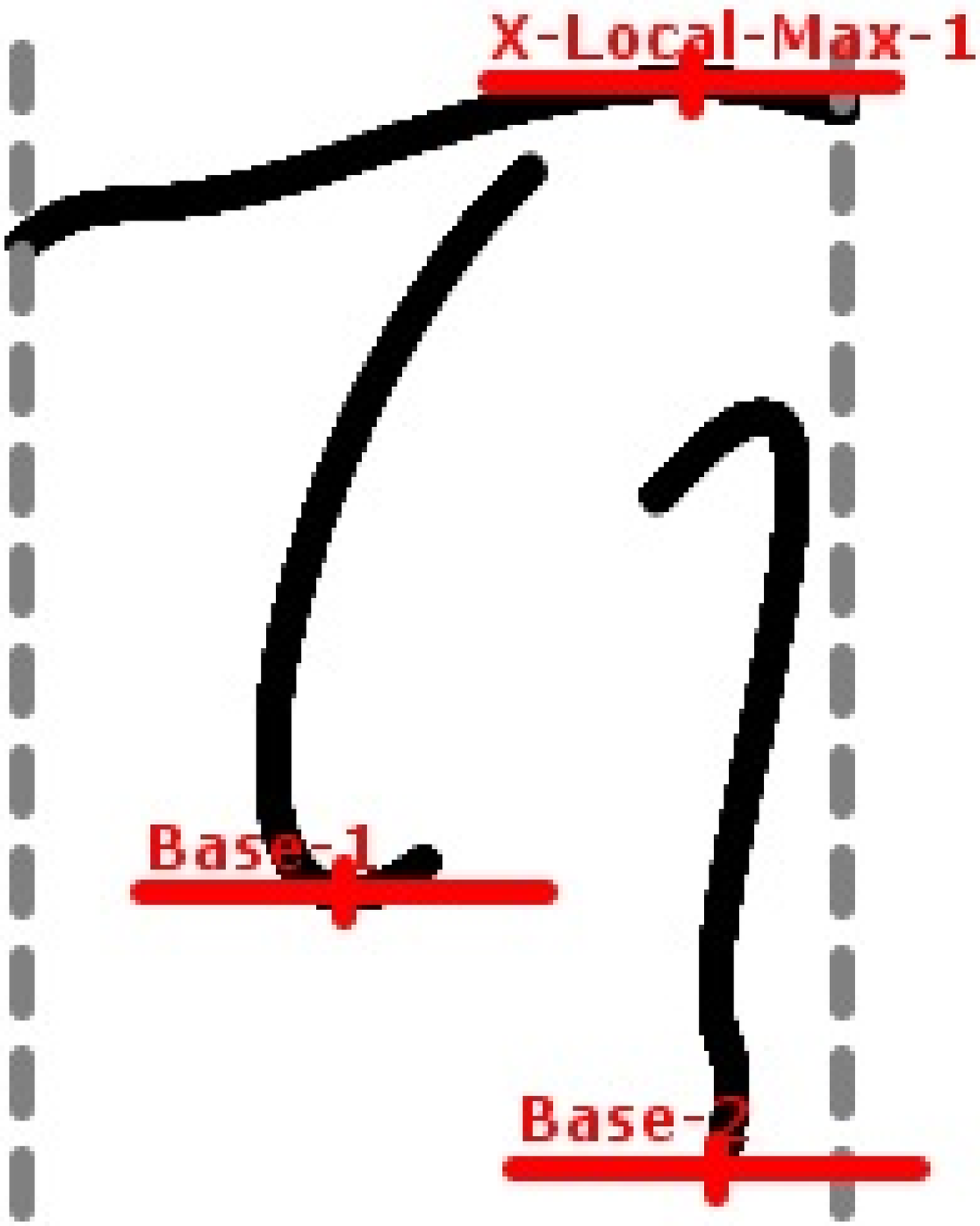}
\label{fig:P4}
}
\subfigure[]{
\includegraphics[width=2.1cm]{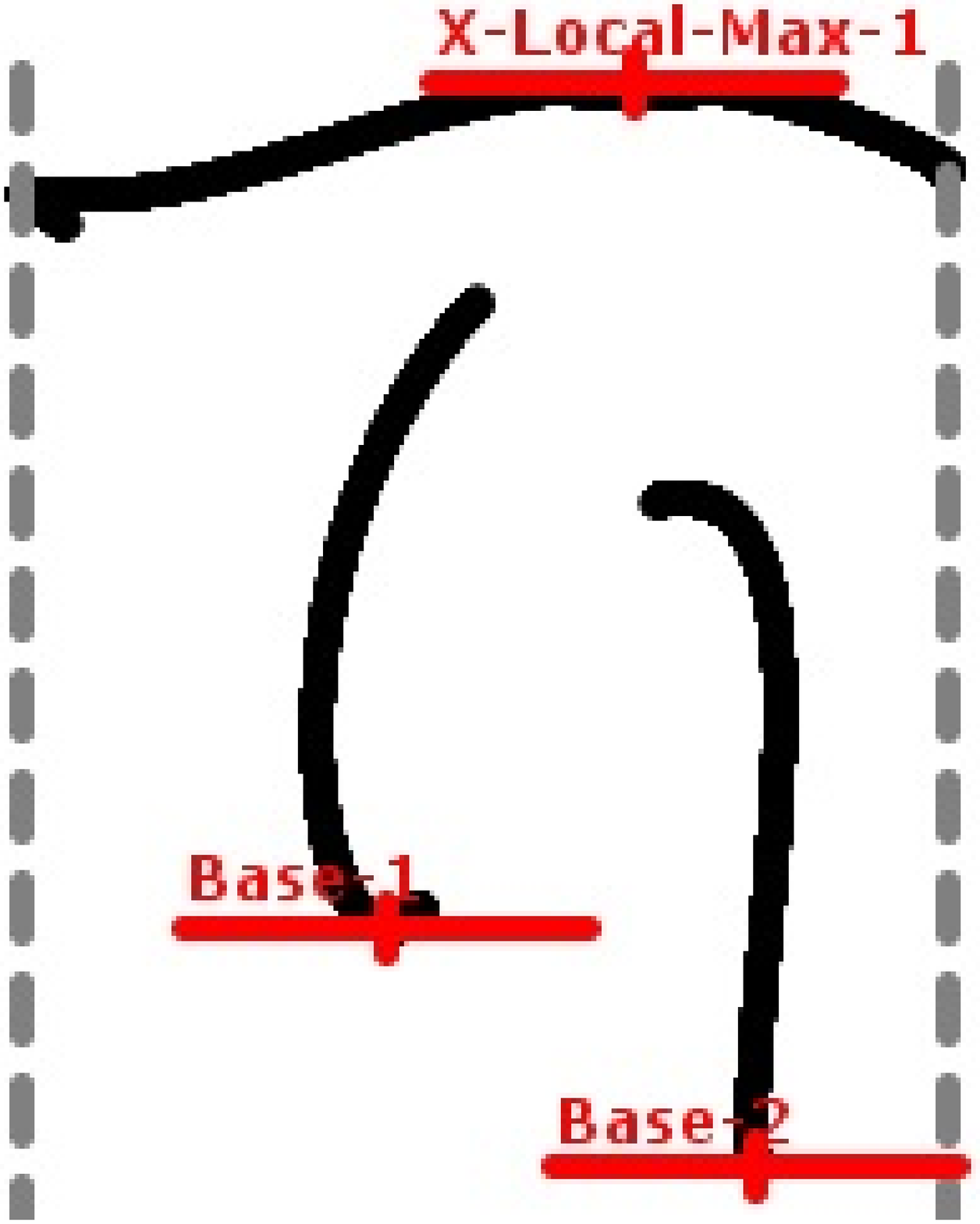}
\label{fig:P5}
}
\subfigure[]{
\includegraphics[width=2.1cm]{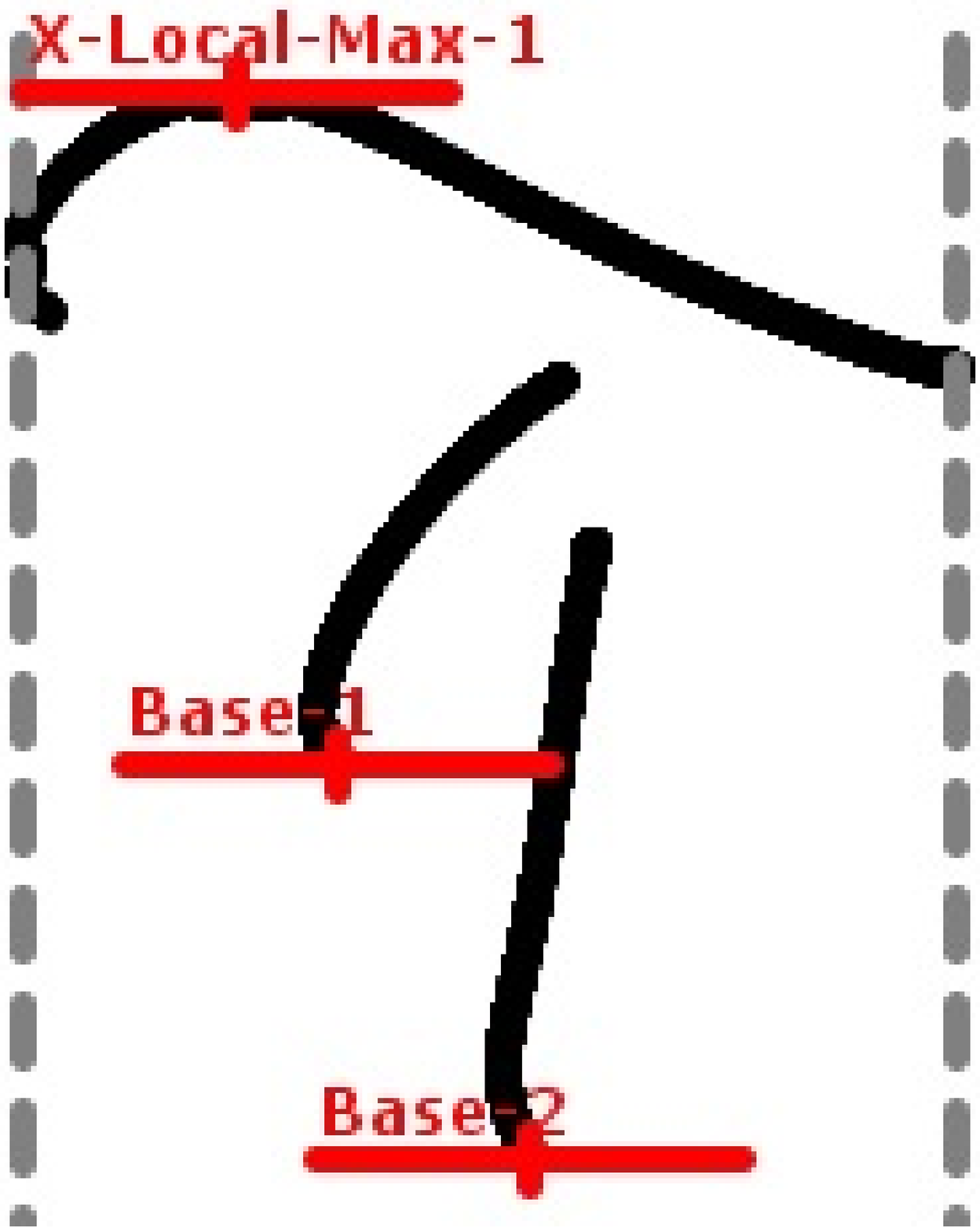}
\label{fig:P3}
}
\subfigure[]{
\includegraphics[width=2.1cm]{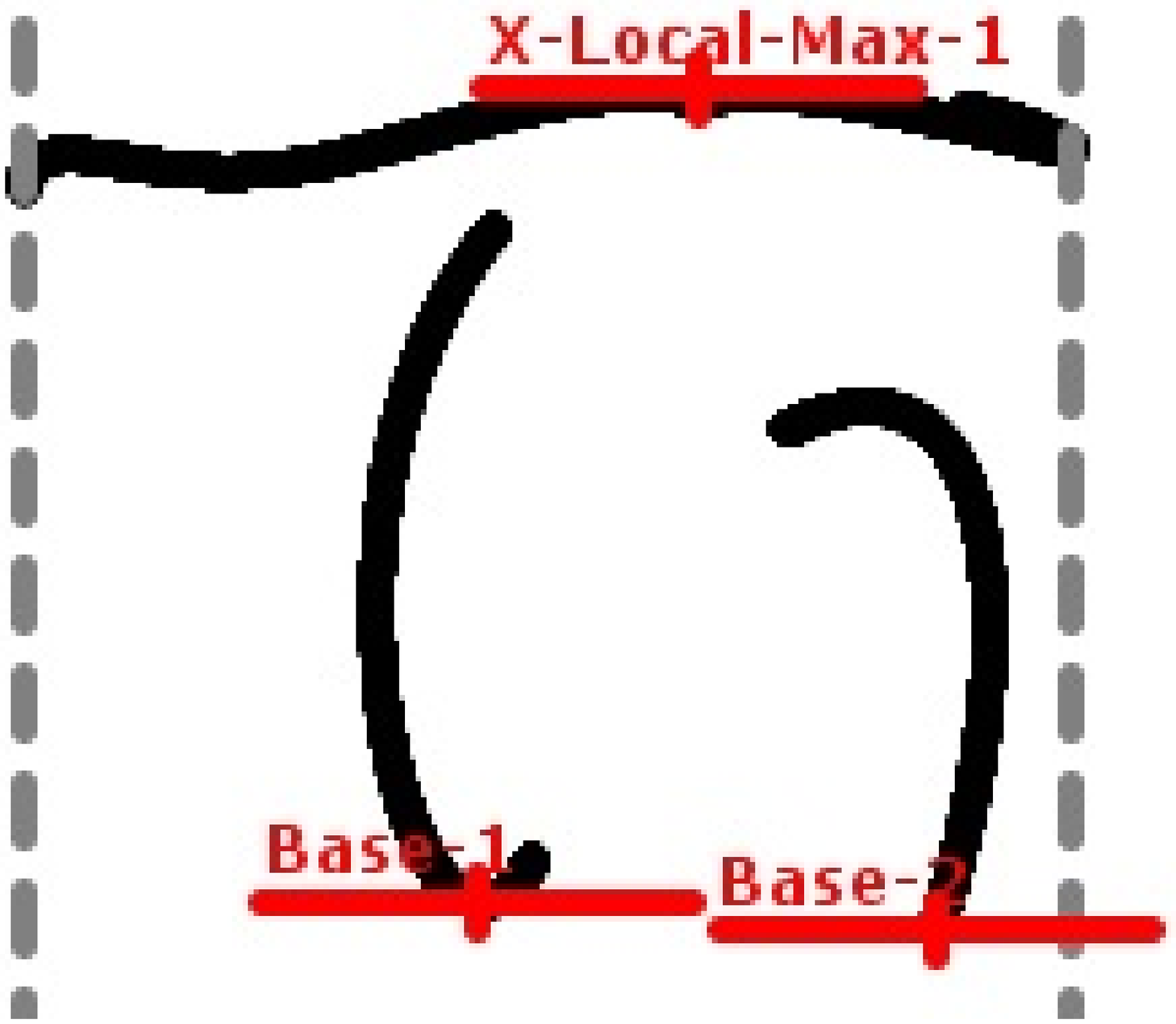}
\label{fig:P1}
}
\caption[]{Automatically finding determining points. 
(a) Average symbol ``$\pi$". \newline (b-e) Determining points derived from the average symbol.}
\label{fig:critical-points-2}
\end{figure}

Figure~\ref{fig:eta-determining-points} shows examples of using Algorithm~\ref{algo:LocateDeterminingPoints}.
Figure~\ref{fig:Average-eta} shows the determining points annotated on the average symbol ``$\eta$''.  This is the reference symbol $A$ in the algorithm.
Figures~\ref{fig:eta-1-initial} and~\ref{fig:eta-2-initial} show two example input samples $S$ with initial approximate locations $s_i$ for the determining points.
Figures~\ref{fig:eta-1-derived} and~\ref{fig:eta-2-derived} show the determining points found at locations $\ell_i$.
Figure~\ref{fig:critical-points-2} shows several examples of determining points found for samples of ``$\pi$''.

\section{Experiments and Testing}\label{sec:experiments}
\begin{figure}[t]
\centering
\setlength{\fboxsep}{0pt}
\setlength{\fboxrule}{2pt}
\fbox{\includegraphics[width=6.5cm]{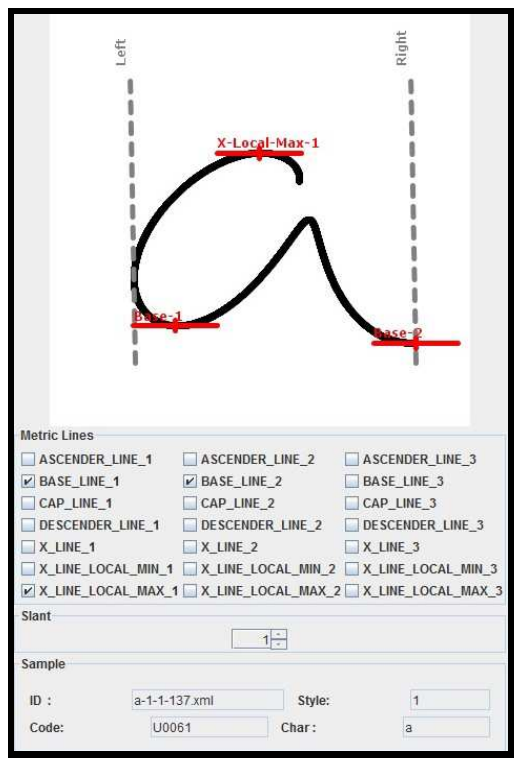}}
\caption{Software tool to identify determining points.}
\label{fig:tool}
\end{figure}

We developed a software tool to annotate handwriting samples with their determining points. Figure
\ref{fig:tool} shows the user interface. By selecting a nearby location,
the tool is able to find the target determining point automatically. The locations of all the metric lines 
 discussed in Section~\ref{sec:metrics} can be detected.
Multiple determining points may exist for certain metrics lines. 
In such circumstances, the location of the corresponding metric line is determined by
the average of the values given by all the determining points of that kind.
Symbol slant can also be recorded by adjusting a spinner. Symbol width is automatically 
detected with slant considered.

To evaluate the performance,
we have tested the algorithm against a large handwriting dataset. The handwriting dataset we used contained altogether 64944 samples of 240
different symbols. Most of the samples are Latin and Greek letters, digits, operators, or other mathematical symbols provided by various writers. All of these samples had been classified in advance. As some symbols were written in different styles 
(e.g. completely different forms, different numbers of strokes, or strokes in different orders), 
a total of 382 classes were examined. We first computed the average symbol for each class, 
in which determining points were identified using the software tool shown in Figure~\ref{fig:tool}. 

We then computed determining points for all the samples using Algorithm~\ref{algo:LocateDeterminingPoints}.
The number of determining points varied from 2 to 5, according to the sample. 
If any of the determining points were mis-positioned, we considered it as incorrect. 
We chose up to 30 samples randomly from each class and examined their correctness visually. 
In total, we examined 8119 samples, of which 421 samples have at least one mis-positioned determining point. 
This gave a measured error rate of 5.2\%.

We found the error was introduced mainly from two sources. The first was mis-classified samples in the original data set.  
These were either mis-labelled (e.g. ``e" of style 1, instead of ``e" of style 2), or had strokes given in a different order from the usual. 
In this latter case, we have the option of defining a new style or normalizing the order of the strokes. 
The second source was that some samples are significantly different from the average symbol. 
As a result, the determining points in the average symbol may not be mapped correctly to those dissimilar samples.

As misclassified samples were errors in the training data,  rather than errors by the algorithm, we excluded those samples from the experiment. We further added 39 new classes (giving 421 classes in total) to split out those samples with different stroke orders. After these corrections, the measured error rate decreased to 2.0\% (9593 samples reviewed, of which 189 samples had at least one mis-positioned determining point).

\begin{figure}[t]
\centering
\subfigure[]{
\includegraphics[width=3.2cm]{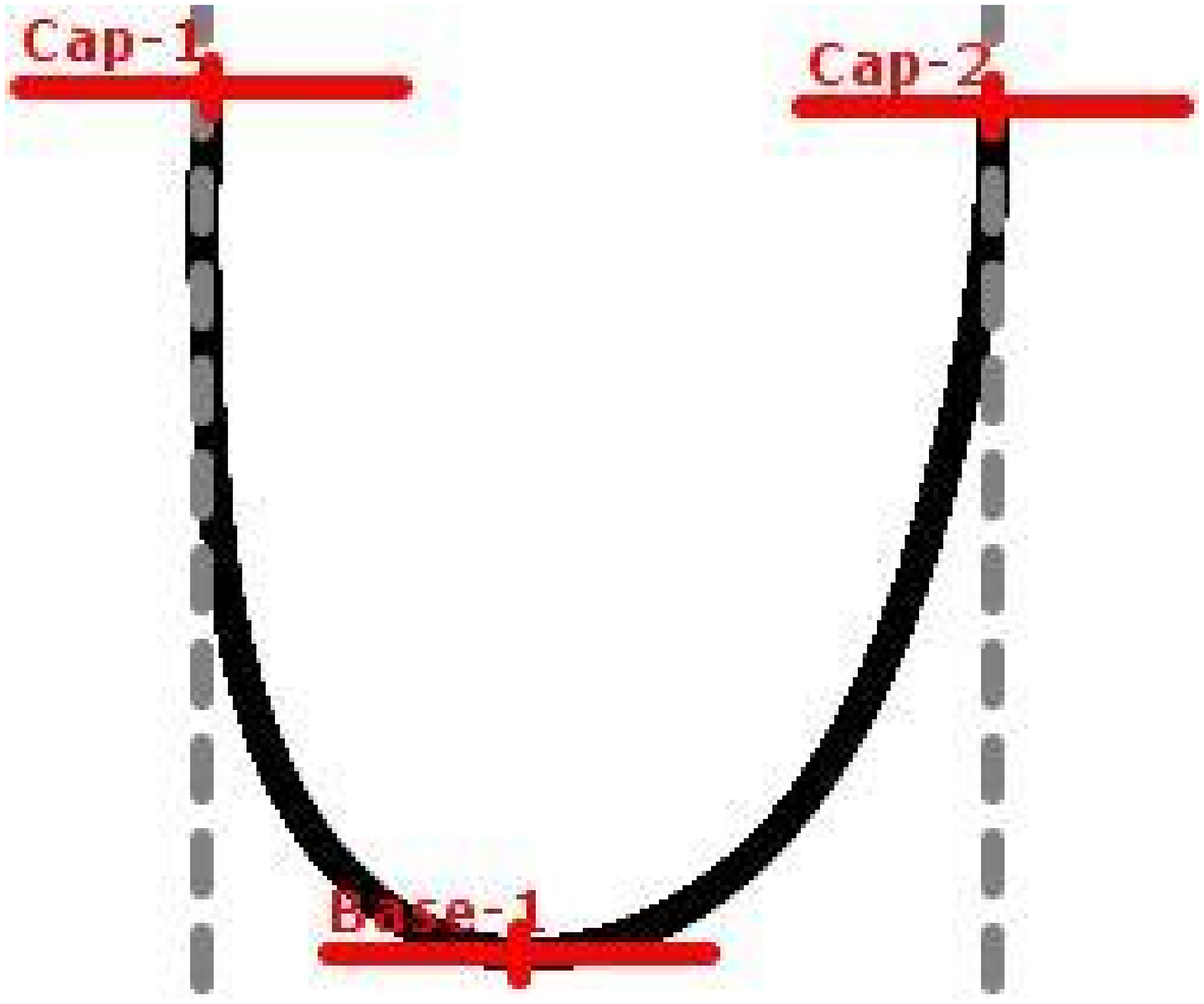}
\label{fig:1-step-average}
}\hspace{5mm}
\subfigure[]{
\includegraphics[width=4.0cm]{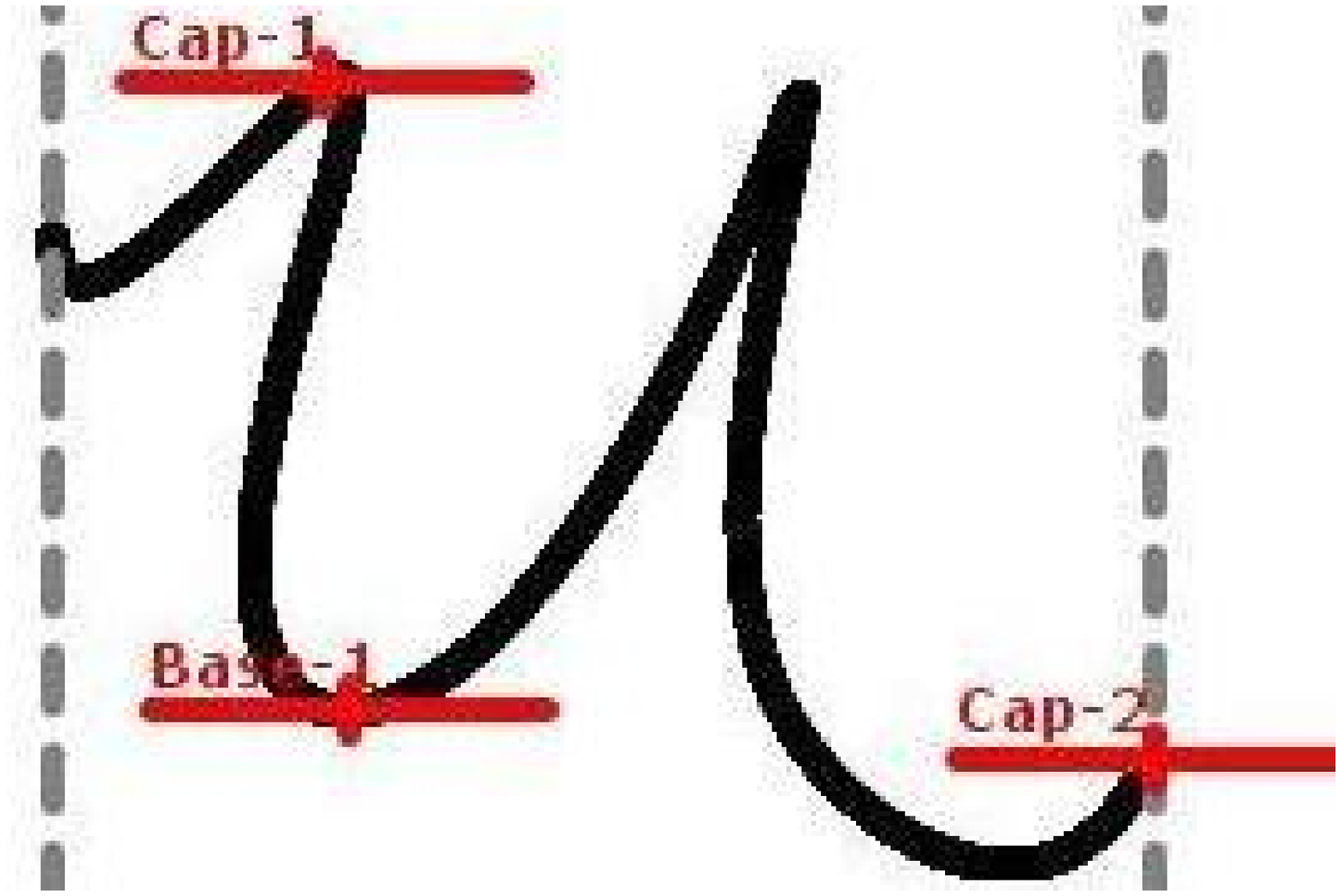}
\label{fig:1-step-sample}
}
\vspace{-.5\baselineskip}
\caption[]{Failure example: (a) average symbol, (b) target with one point misplaced.}
\label{fig:1-step-algorithm}
\end{figure}

\begin{figure}[t]
\centering
\subfigure[]{
\includegraphics[width=2.55cm]{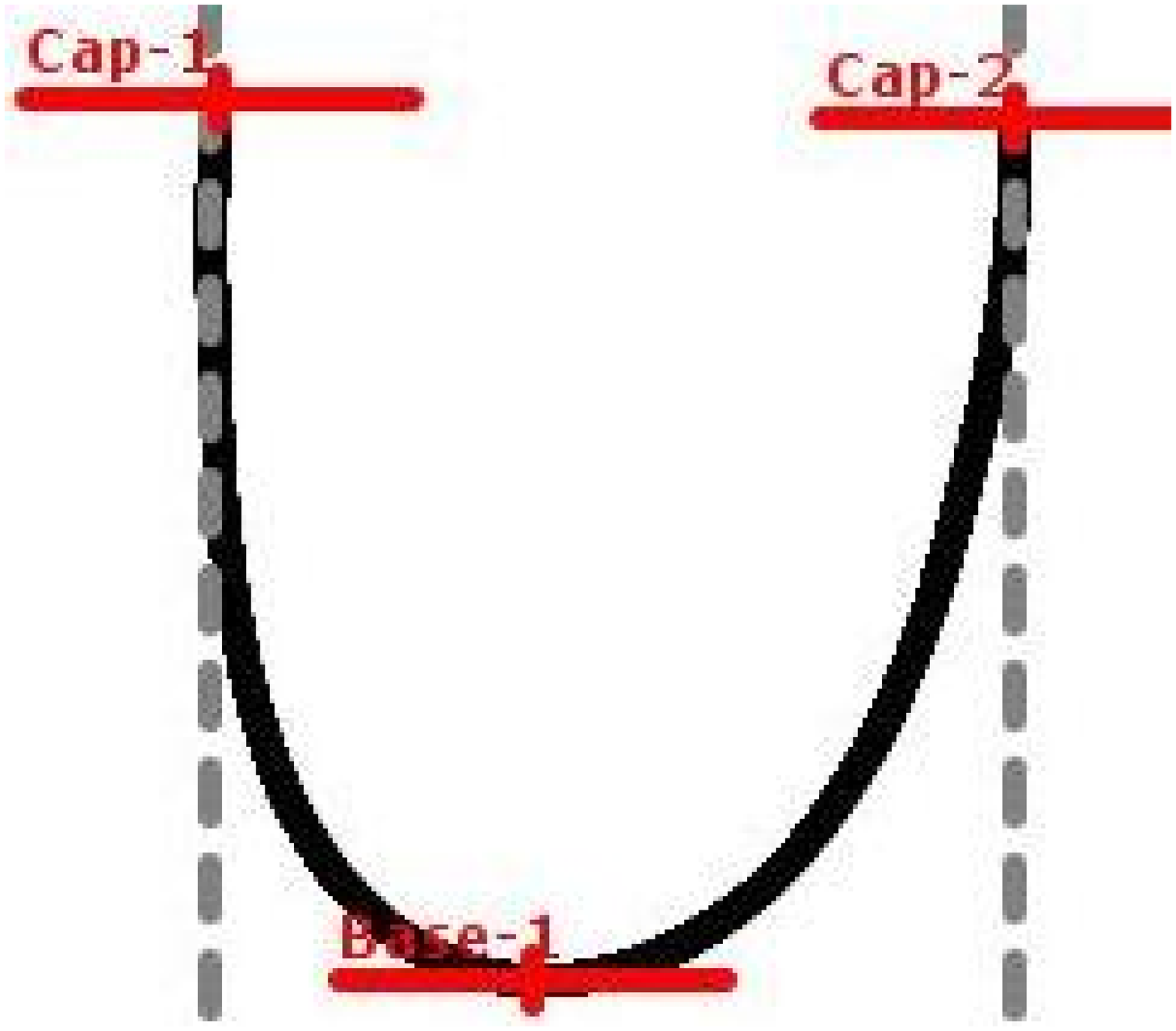}
\label{fig:3-step-average}
}\hspace{-3mm}
\subfigure[]{
\includegraphics[width=3.2cm]{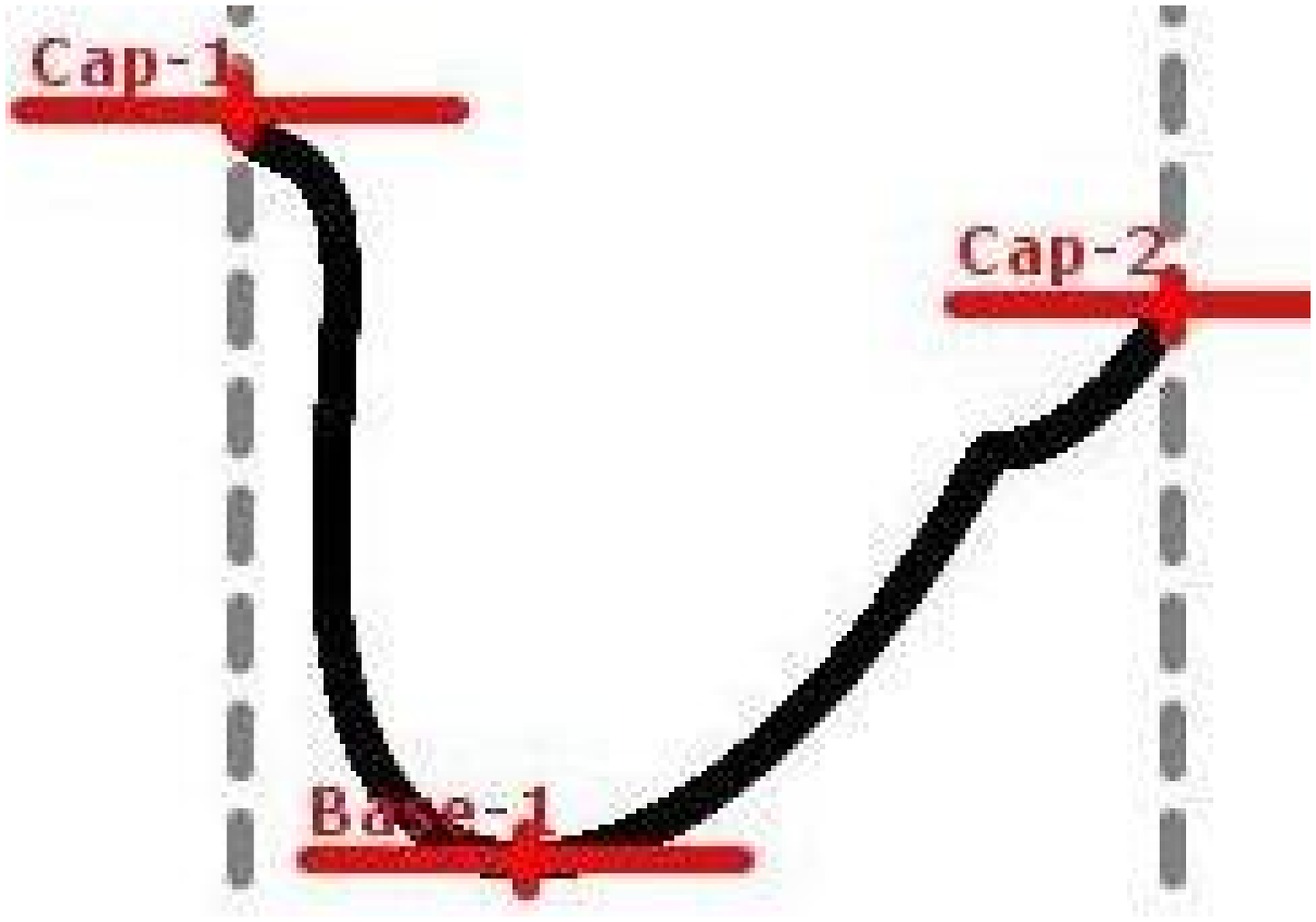}
\label{fig:3-step-1}
}\hspace{-3mm}
\subfigure[]{
\includegraphics[width=3.1cm]{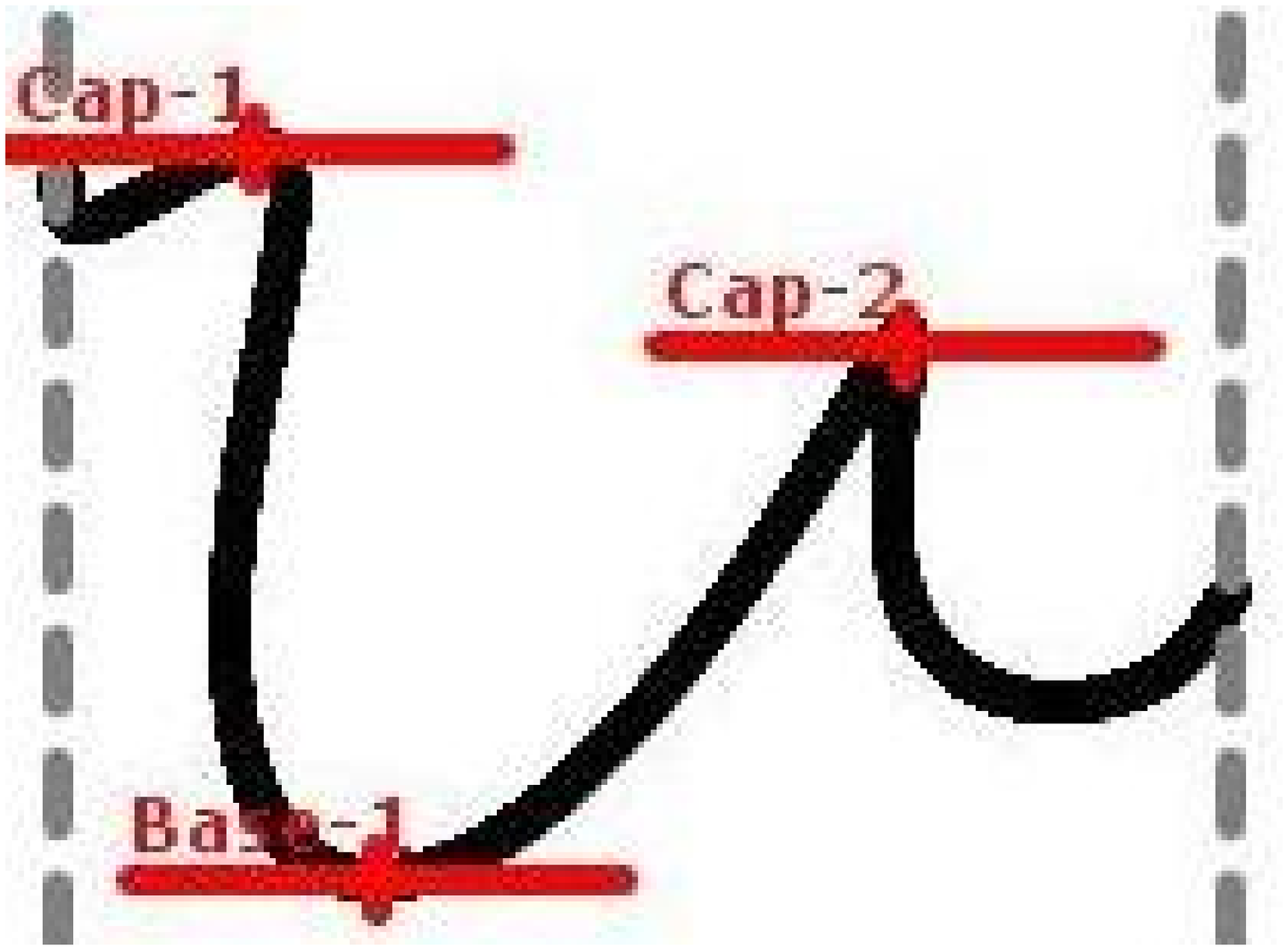}
\label{fig:3-step-2}
}\hspace{-3mm}
\subfigure[]{
\includegraphics[width=3.0cm]{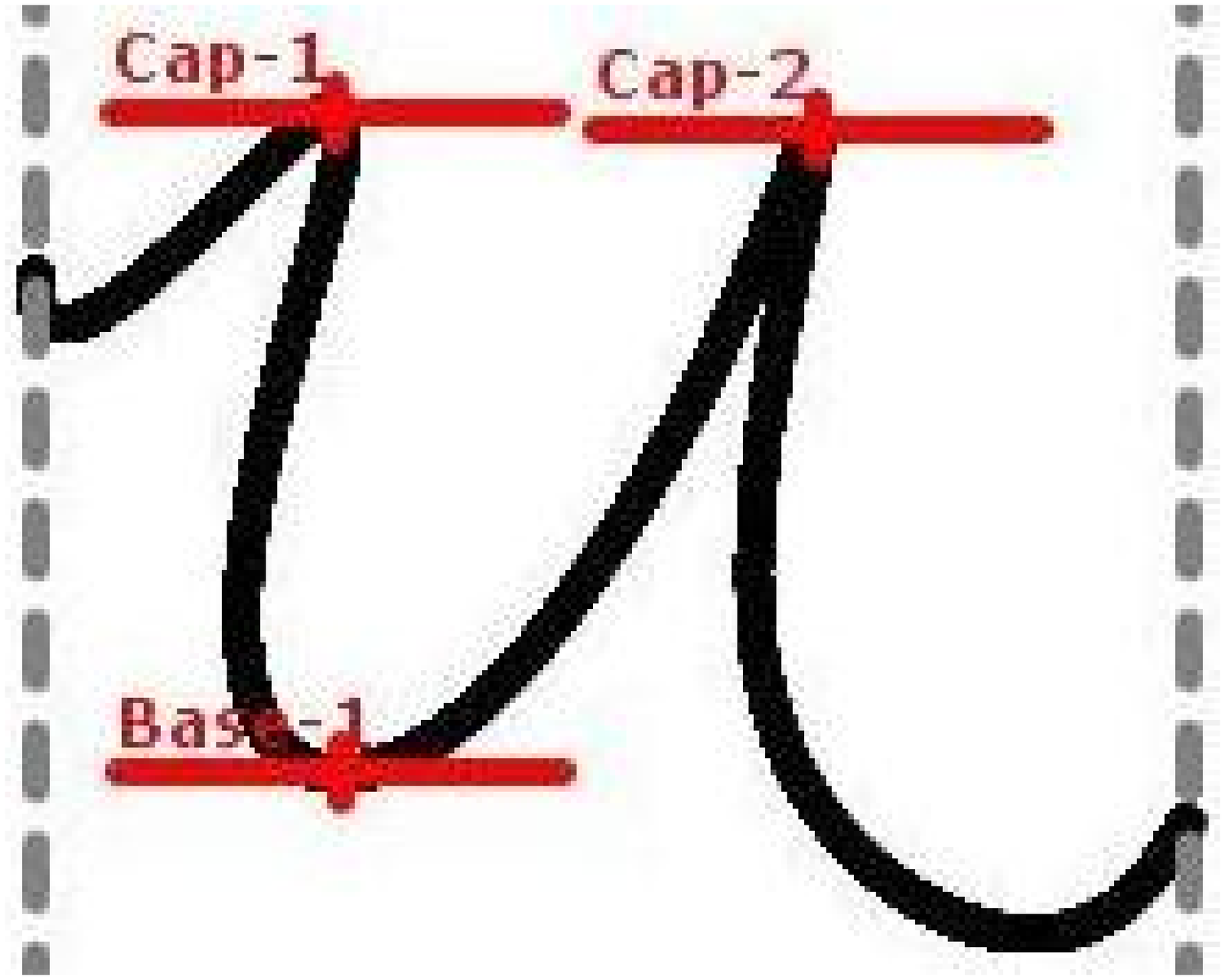}
\label{fig:3-step-sample}
}
\vspace{-.5\baselineskip}
\caption[]{Success in 3 steps: (a) average (b) step 1 (c) step 2 (d) step 3 = target.}
\label{fig:multi-step-algorithm}
\end{figure}

To address the second issue, that of points mis-positioned because the sample was far from the average shape, 
we used a homotopy between the average and the test sample in a multi-step method. Recall that, in the function space, a line from the average symbol to the test sample lies entirely within the class. By dividing this line into several equal steps, we may apply Algorithm
\ref{algo:LocateDeterminingPoints} several times to follow the determining points through the homotopy.
If  $\bar C$ is the average symbol for the class and $C_{targ}$ is the input sample, then the line joining the two points in the function space is given by $C(t) = (1-t) \bar C + t C_{targ}$, with $t$ ranging from 0 to 1. The determining points should move smoothly as the character is deformed by the homotopy, and we can choose a step size. 
Figure~\ref{fig:1-step-algorithm} shows an example where Algorithm~\ref{algo:LocateDeterminingPoints} fails to identify one of the determining points when applied naively.
However, when applied in a 3 step homotopy, it succeeded, as shown in Figure~\ref{fig:multi-step-algorithm}.

\begin{figure}[t]
\centering
\subfigure[]{
\includegraphics[width=3.0cm]{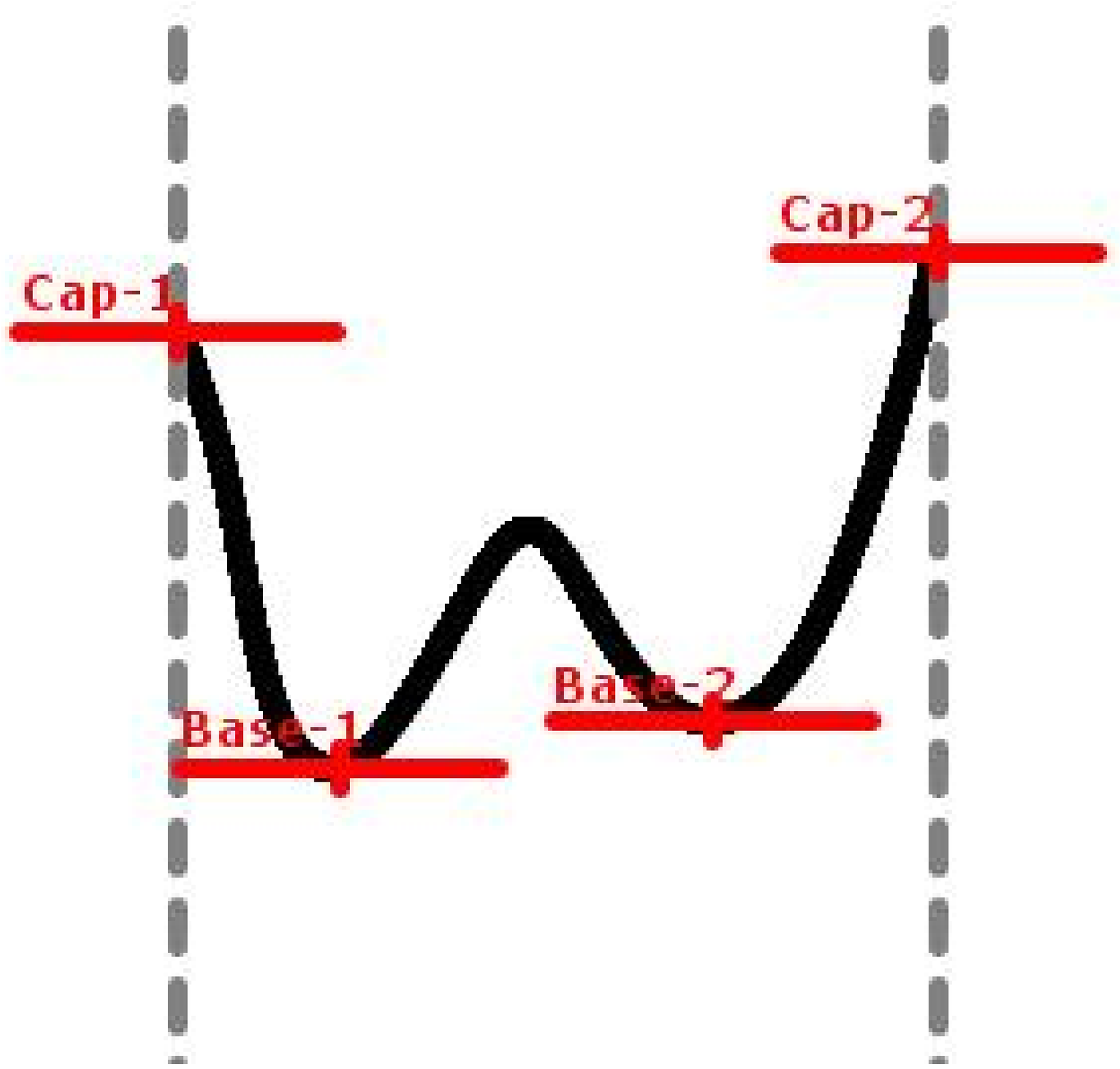}
\label{fig:W-average}
}\hspace{-5mm}
\subfigure[]{
\includegraphics[width=3.0cm]{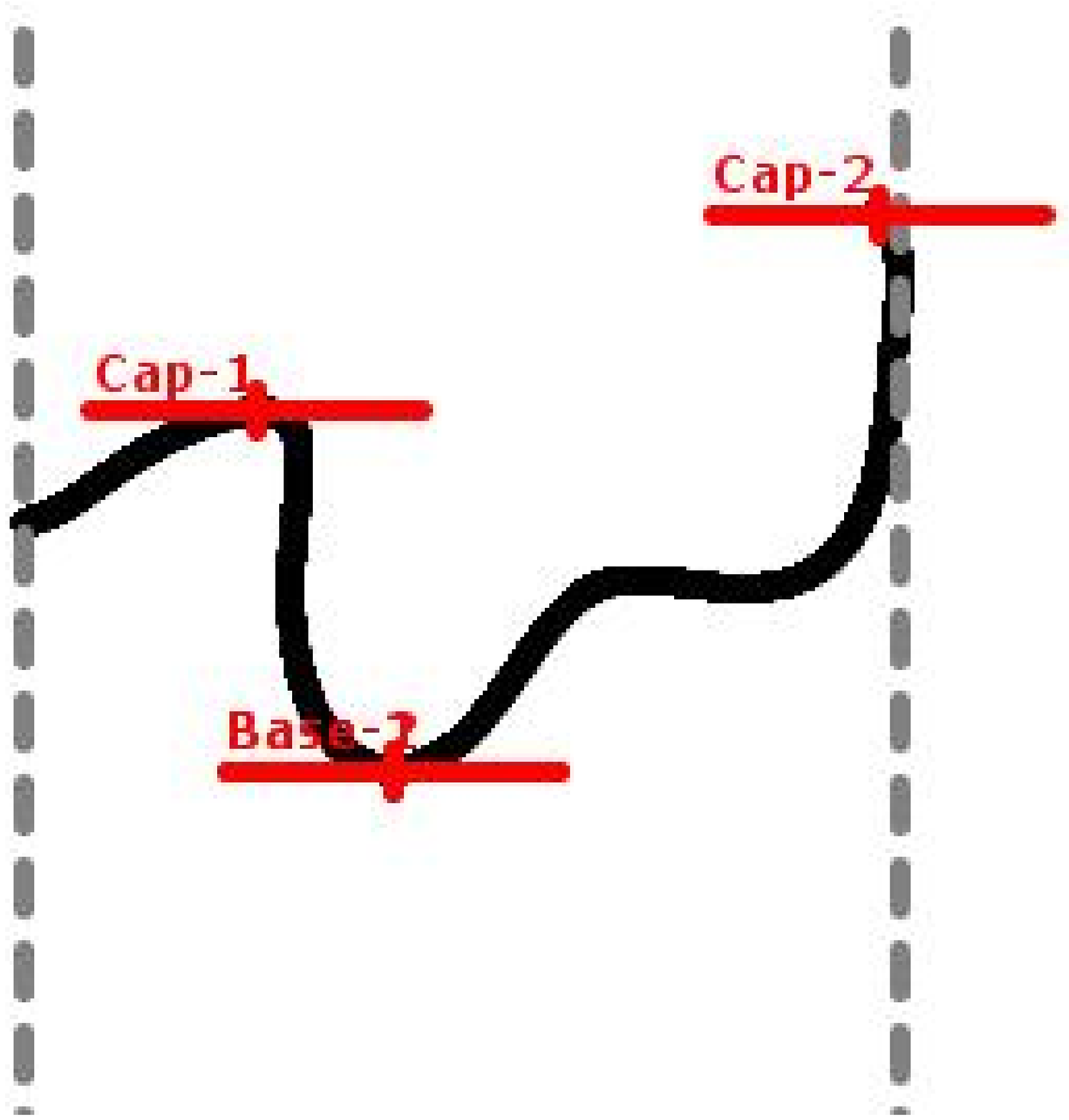}
\label{fig:W-bad}
}\subfigure[]{
\includegraphics[width=3.0cm]{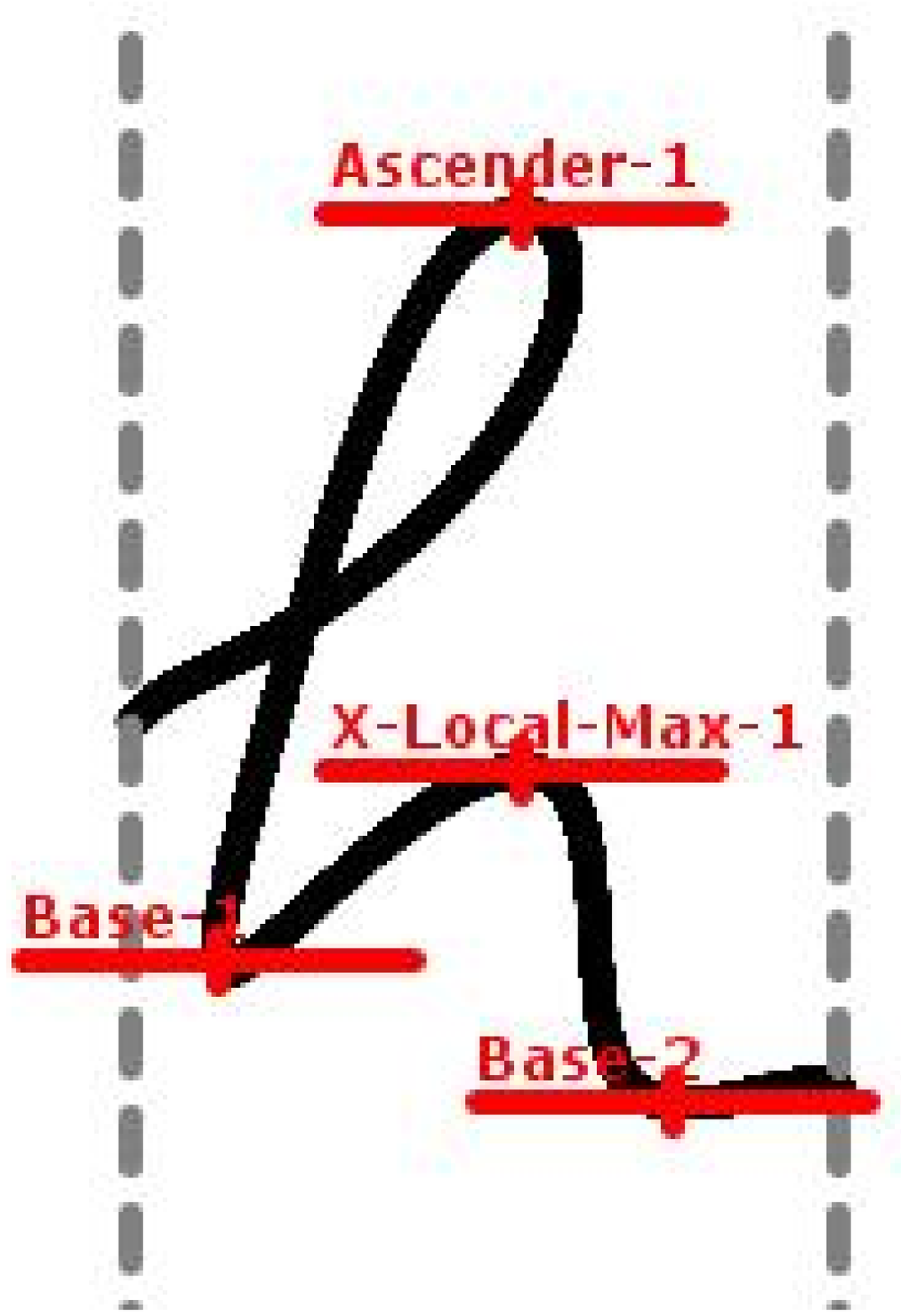}
\label{fig:h-average}
}\hspace{-5mm}
\subfigure[]{
\includegraphics[width=3.0cm]{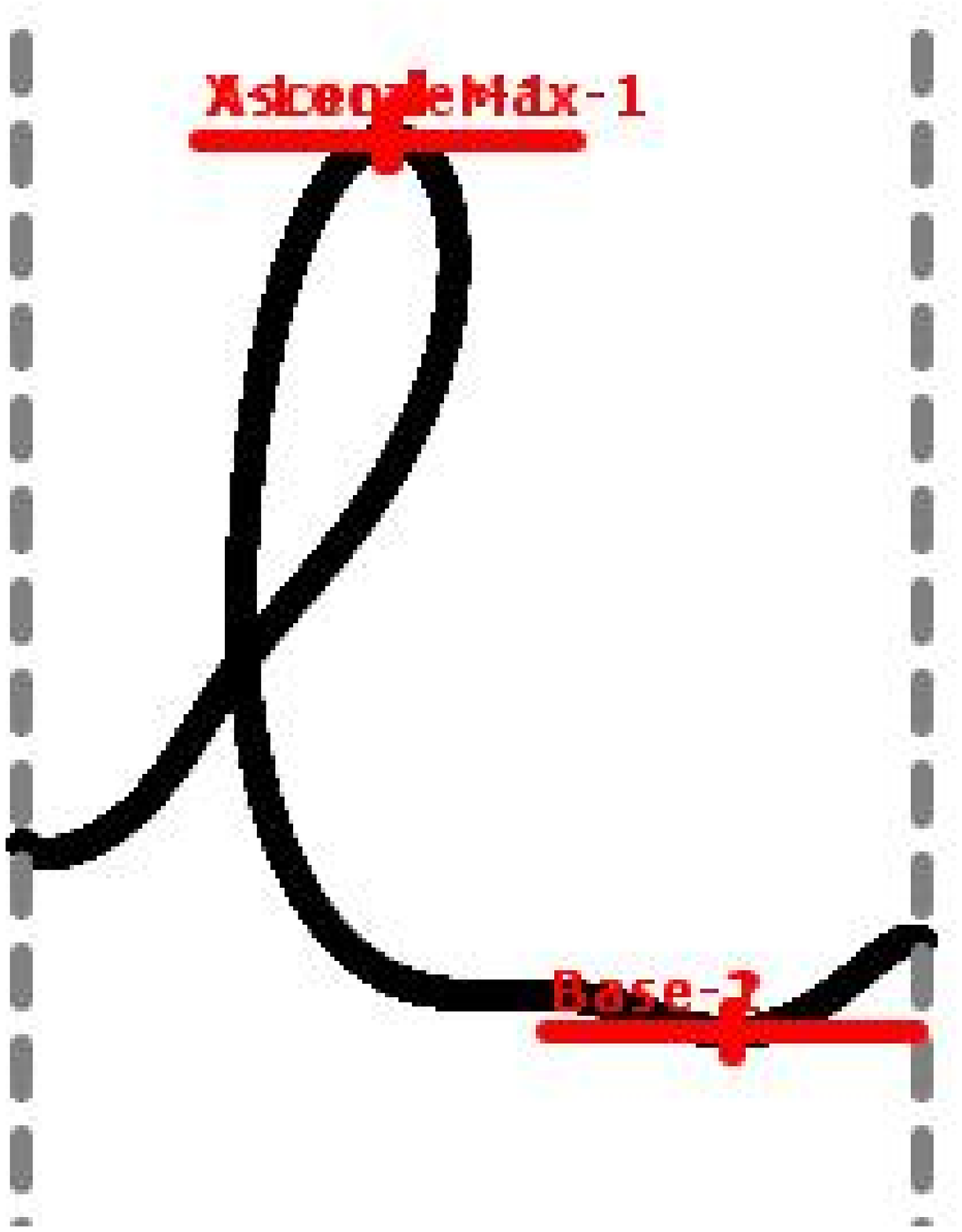}
\label{fig:h-bad}
}
\caption[]{Multi-step failures: (a) Average, (b) target. (c) Average, (d) target.}
\label{fig:bad}
\end{figure}

\begin{table}[t]
\begin{center}
\vspace{.5\baselineskip}
    \begin{tabular}{ | c | c | c | c | c | c | c | c | c |}
    \hline
    \rule{0pt}{10pt} Steps & 1 & 2 & 3 & 4 & 6 & 8 & 10 & 20 \\ \hline
    \rule{0pt}{10pt} Failed Samples & 189 & 69 & 36 & 28 & 25 & 25 & 24 & 24 \\ \hline
    \rule{0pt}{10pt} Error Rates & 2.0\% & 0.72\% & 0.38\% & 0.29\% & 0.26\% & 0.26\% & 0.25\% & 0.25\% \\ \hline
    \end{tabular}
\end{center}
\caption{Error rates of the multi-step method on 9593 samples.}
\label{tab:error-rates}
\end{table}
\enlargethispage{.5\baselineskip}
We have tested the multi-step method against the same handwriting dataset.  We chose up to 30 samples randomly from each class and examined their correctness visually. The measured error rates  are reported in Table~\ref{tab:error-rates}.
The samples that failed in the 10-Step and 20-Step methods typically either had slants that interfered with the strategy of using local minimum or maximum $y$ value to find determining points or that were very badly written. For these samples, our algorithm was able to identify some determining points correctly but not all of them, as shown in Figure~\ref{fig:bad}.   Note that the points found would in any case be sufficient for most applications.

\section{Use Cases}\label{sec:use-cases}
Determining points can be used in a variety of digital ink applications to solve different problems.
Here we describe two scenarios in which determining points have been found useful.

\subsection*{Handwriting Recognition}

\begin{figure}[t]
\centering
\subfigure[]{
\includegraphics[width=1.4cm]{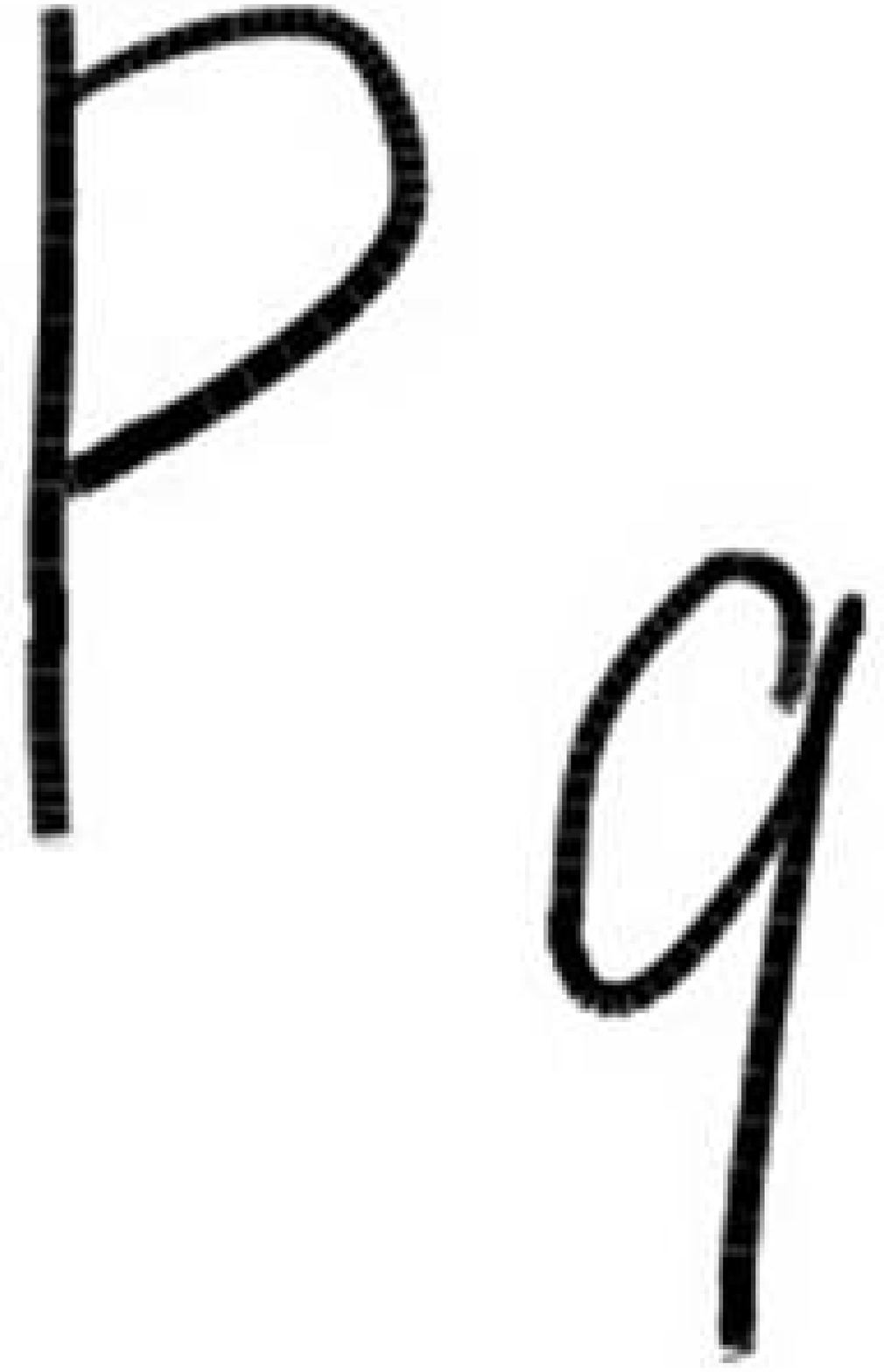}
\label{fig:pq1}
}\hspace{3mm}
\subfigure[]{
\includegraphics[width=1.4cm]{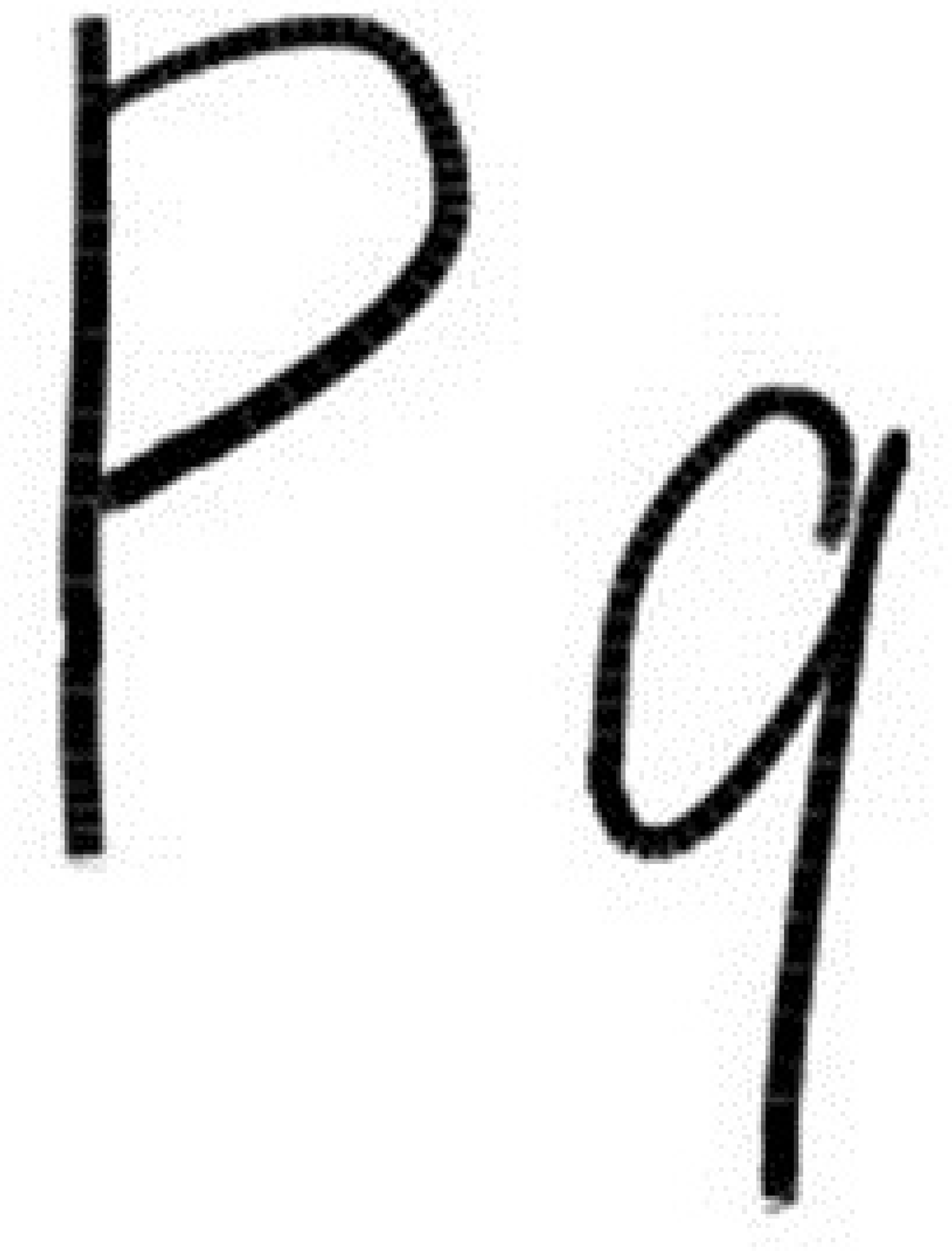}
\label{fig:pq2}
}\hspace{3mm}
\subfigure[]{
\includegraphics[width=1.4cm]{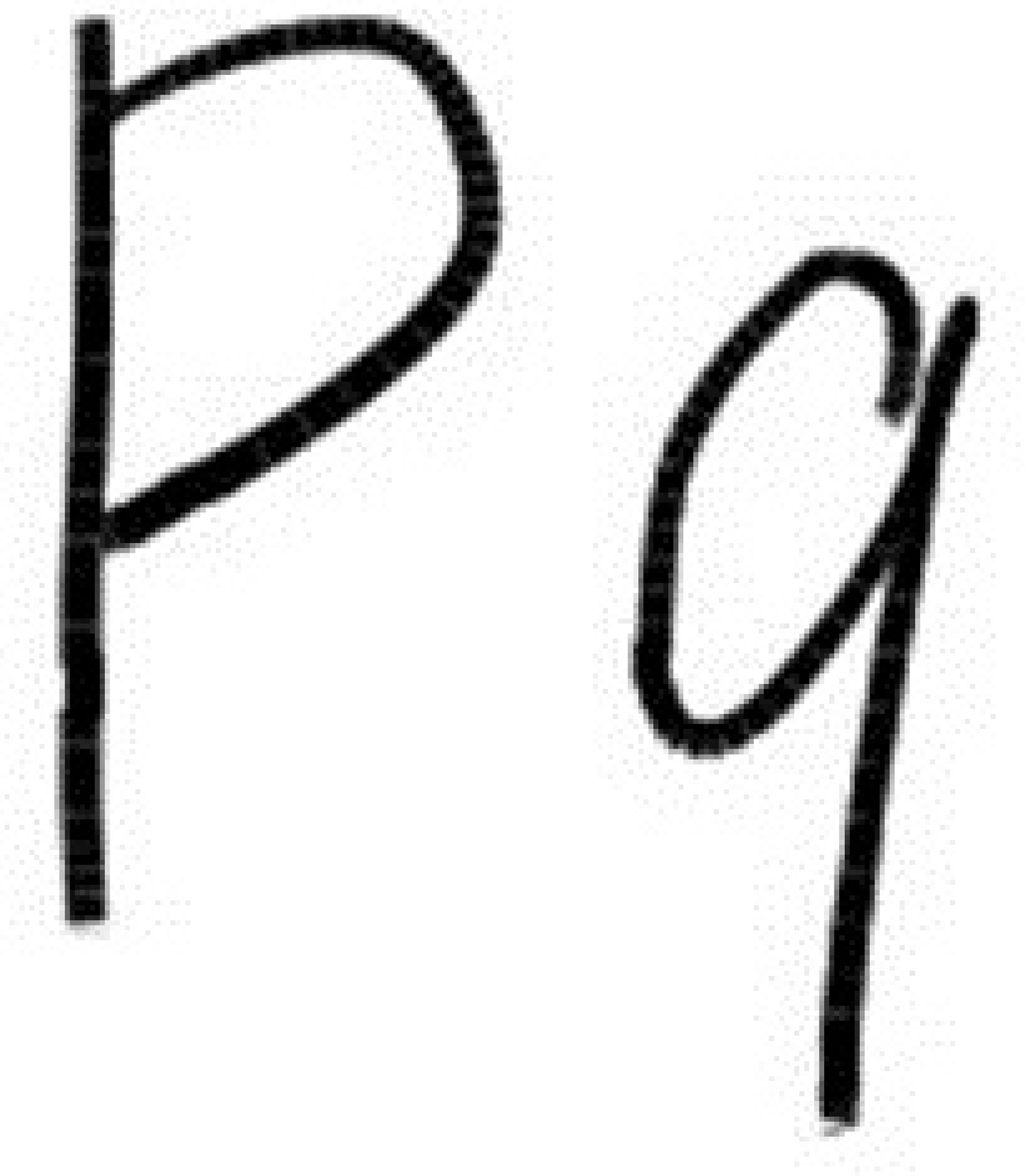}
\label{fig:pq3}
}\hspace{3mm}
\subfigure[]{
\includegraphics[width=1.4cm]{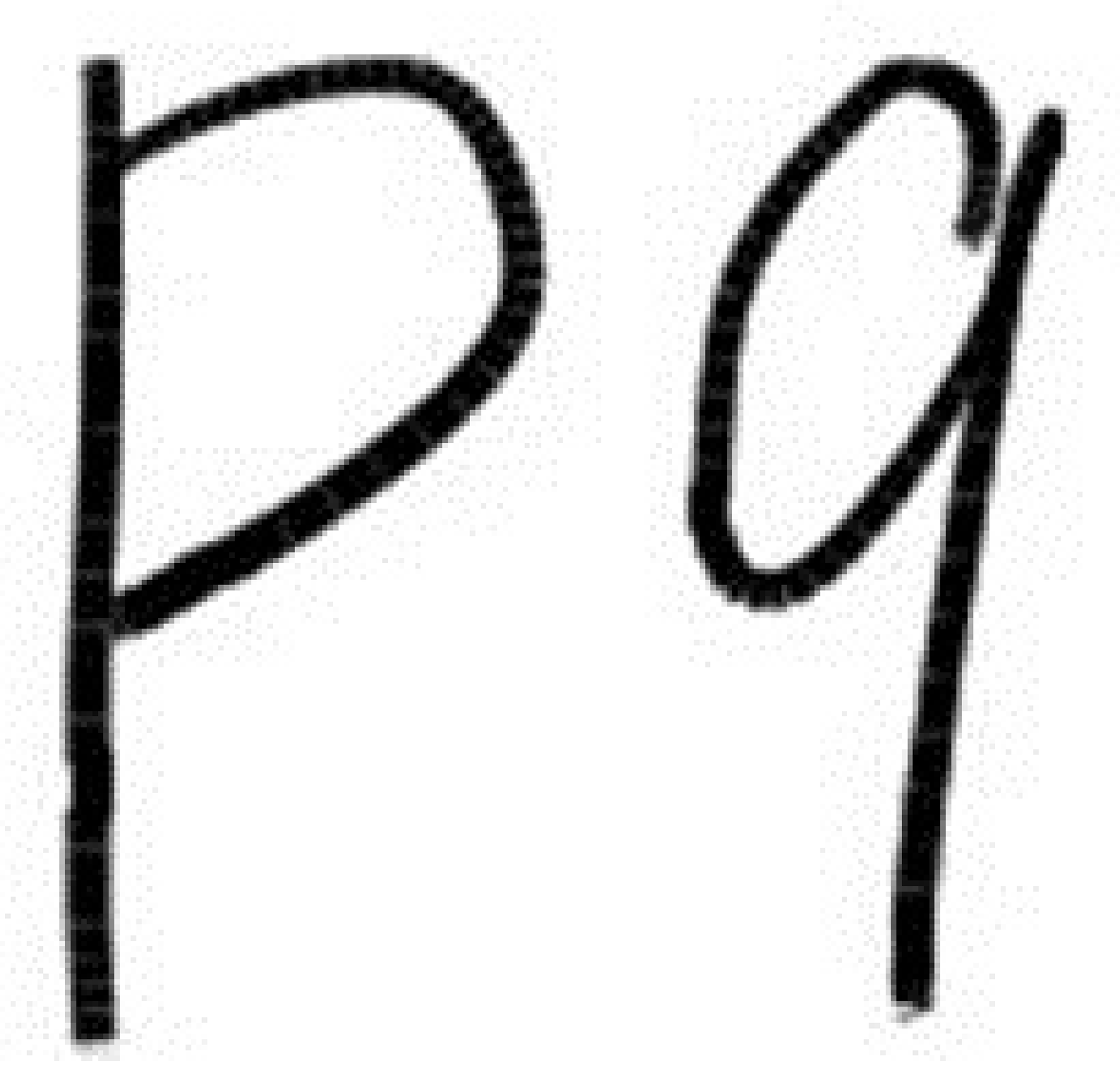}
\label{fig:pq4}
}\hspace{3mm}
\subfigure[]{
\includegraphics[width=1.4cm]{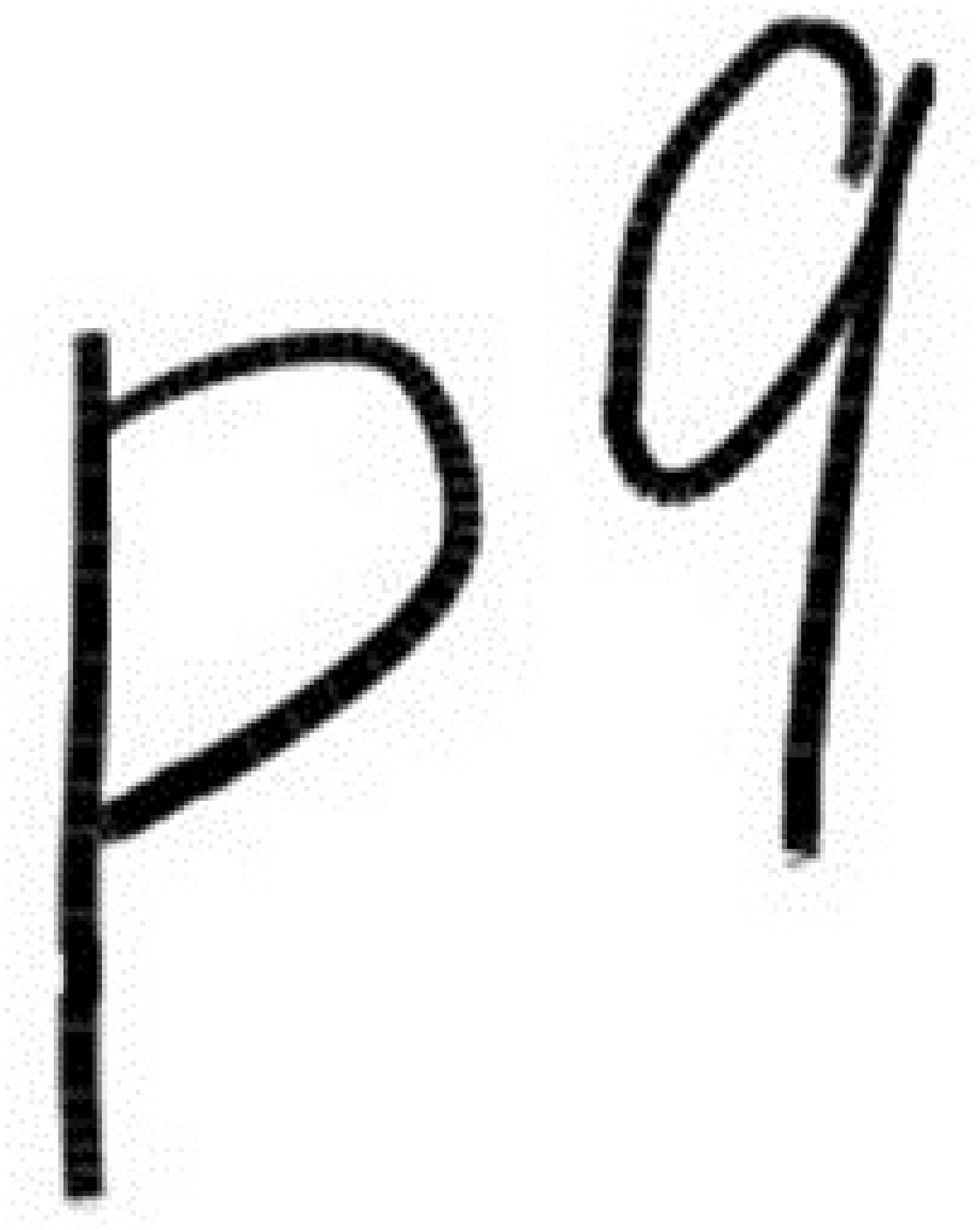}
\label{fig:pq5}
}\hspace{3mm}
\subfigure[]{
\includegraphics[width=1.4cm]{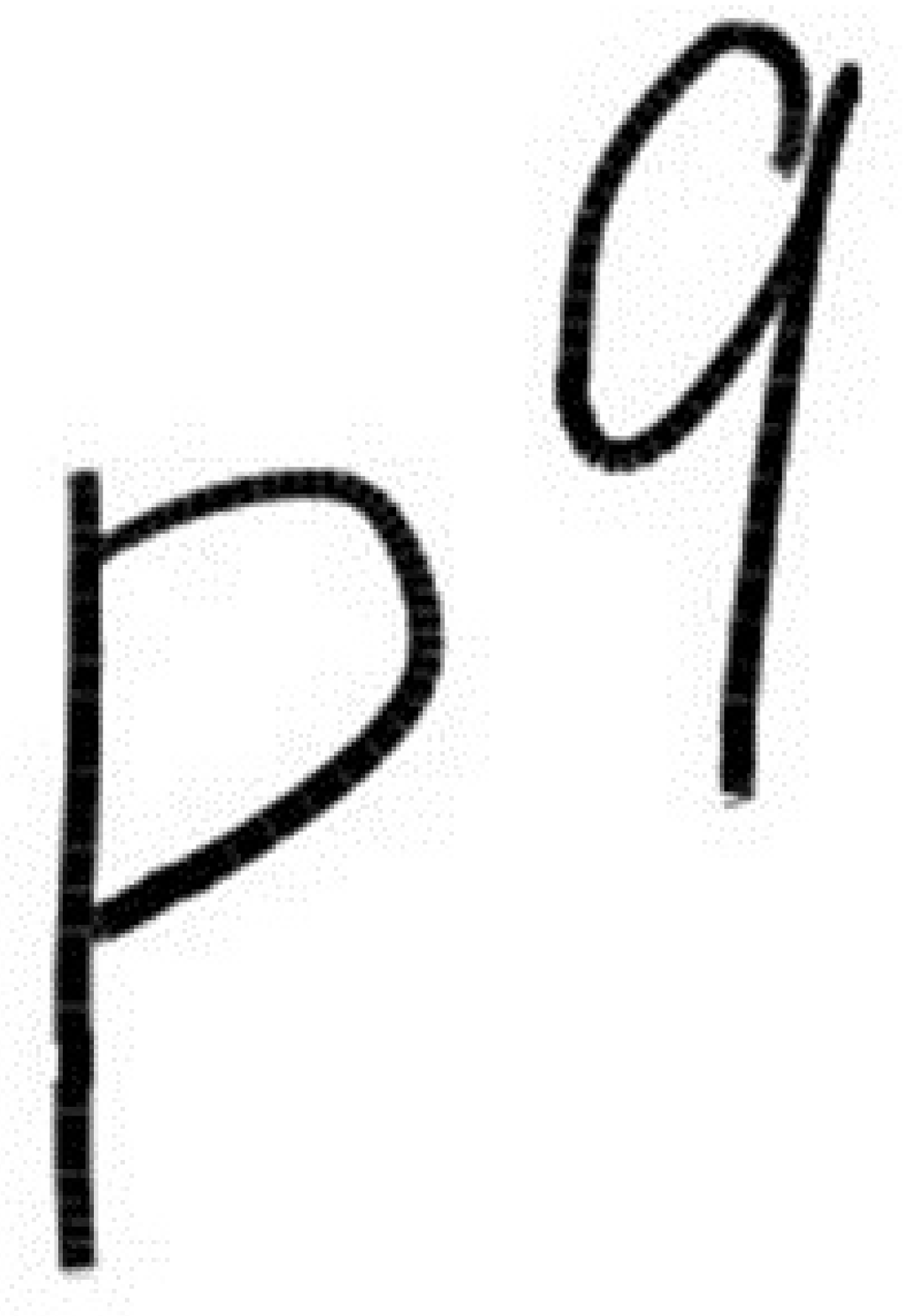}
\label{fig:pq6}
}
\caption[]{Juxtaposition ambiguity.}
\label{fig:juxtaposition}
\end{figure}

\begin{figure}[t]
\centering
\subfigure[]{
\includegraphics[width=1.8cm]{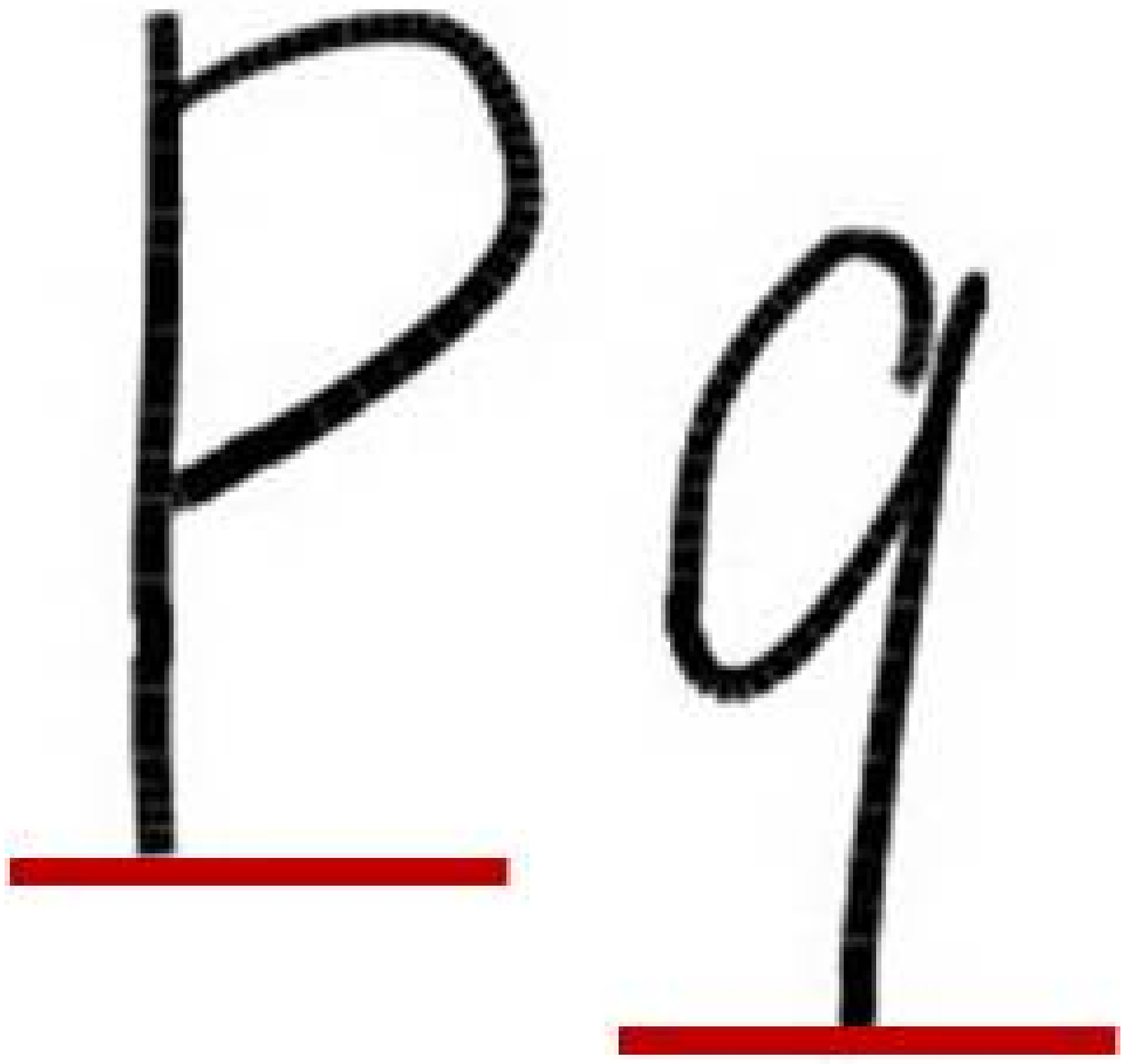}
\label{fig:P9}
}\hspace{3mm}
\subfigure[]{
\includegraphics[width=1.8cm]{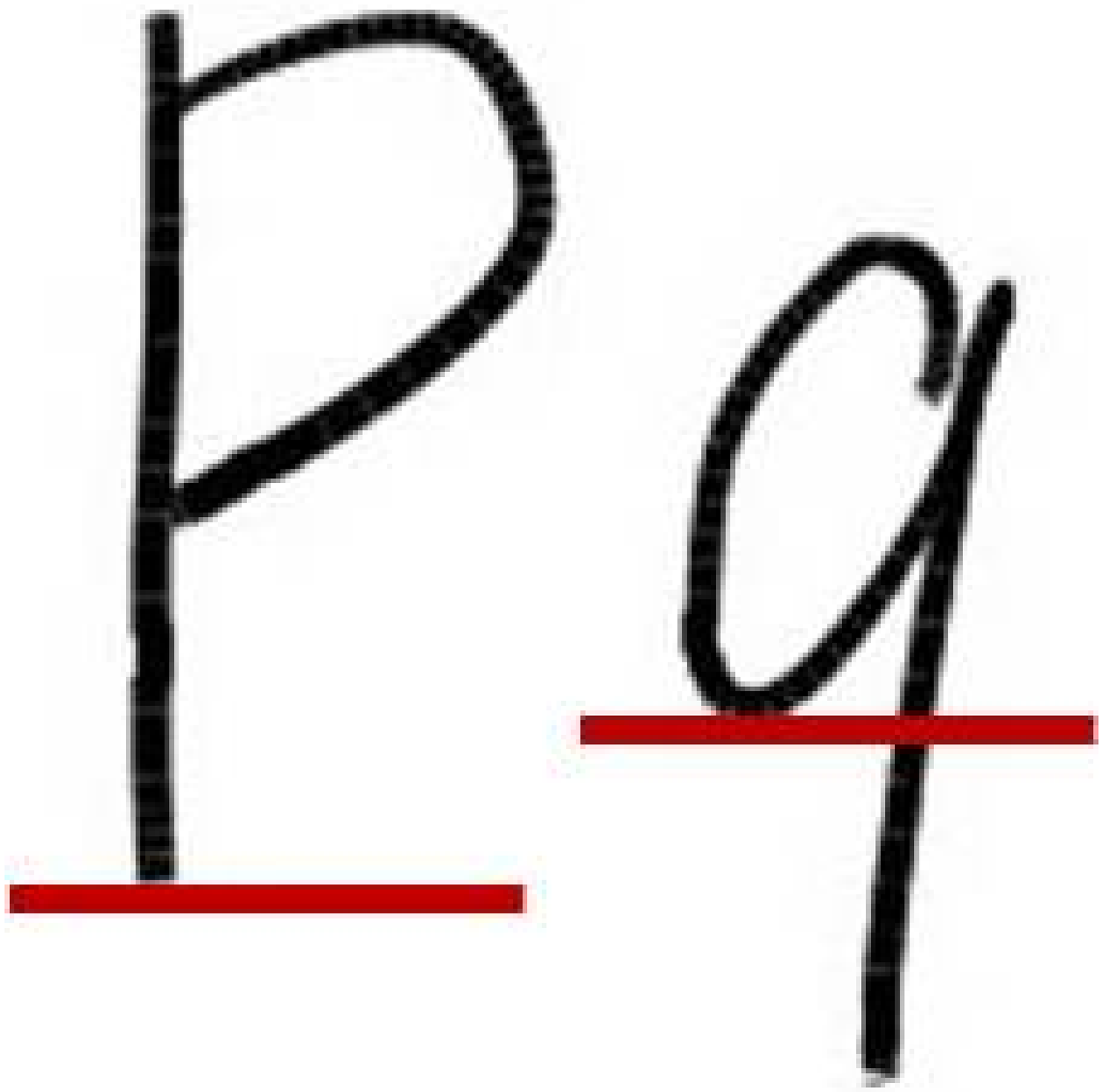}
\label{fig:Pq}
}\hspace{3mm}
\subfigure[]{
\includegraphics[width=1.8cm]{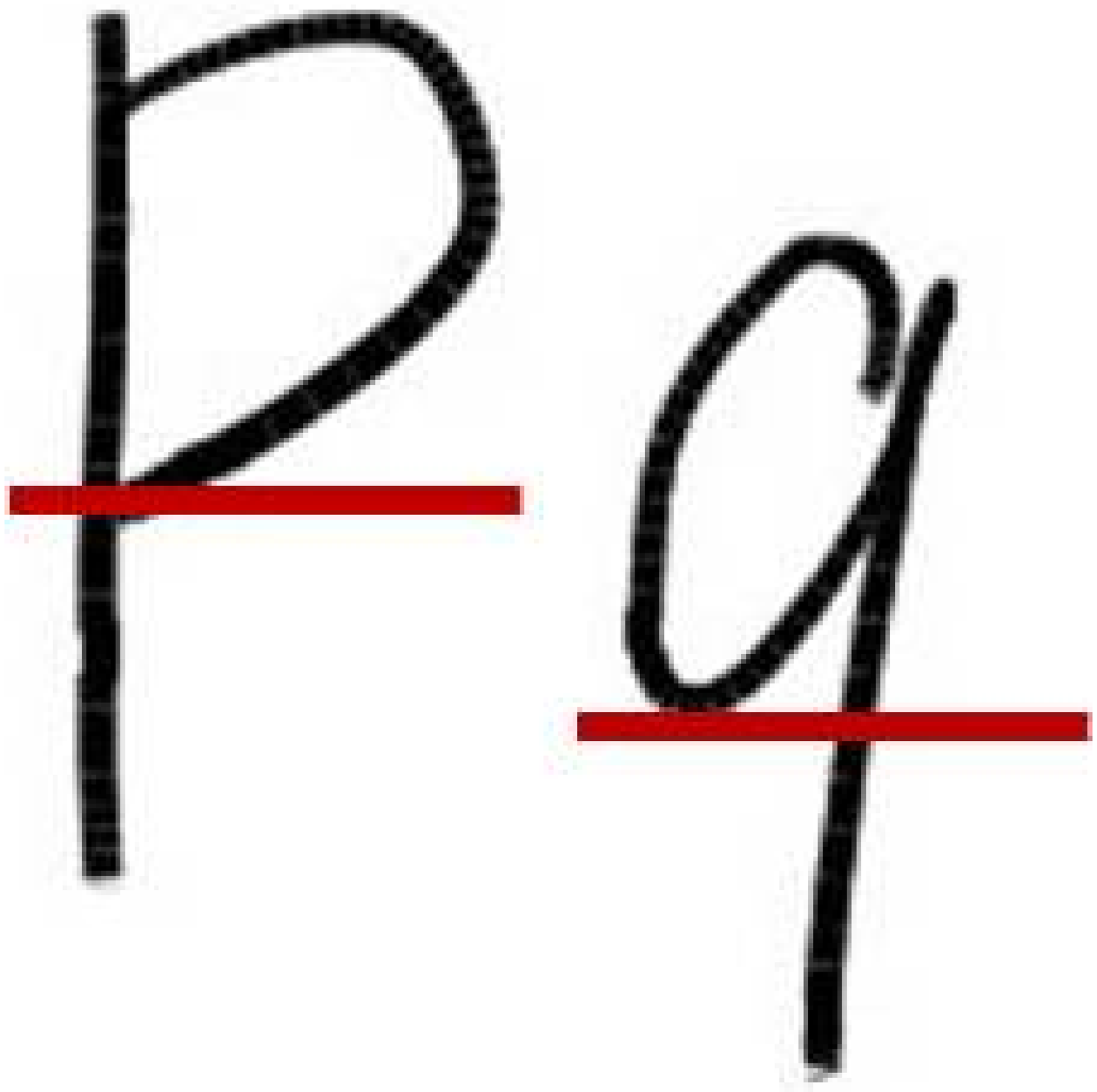}
\label{fig:pq-lowercase}
}\hspace{3mm}
\subfigure[]{
\includegraphics[width=1.8cm]{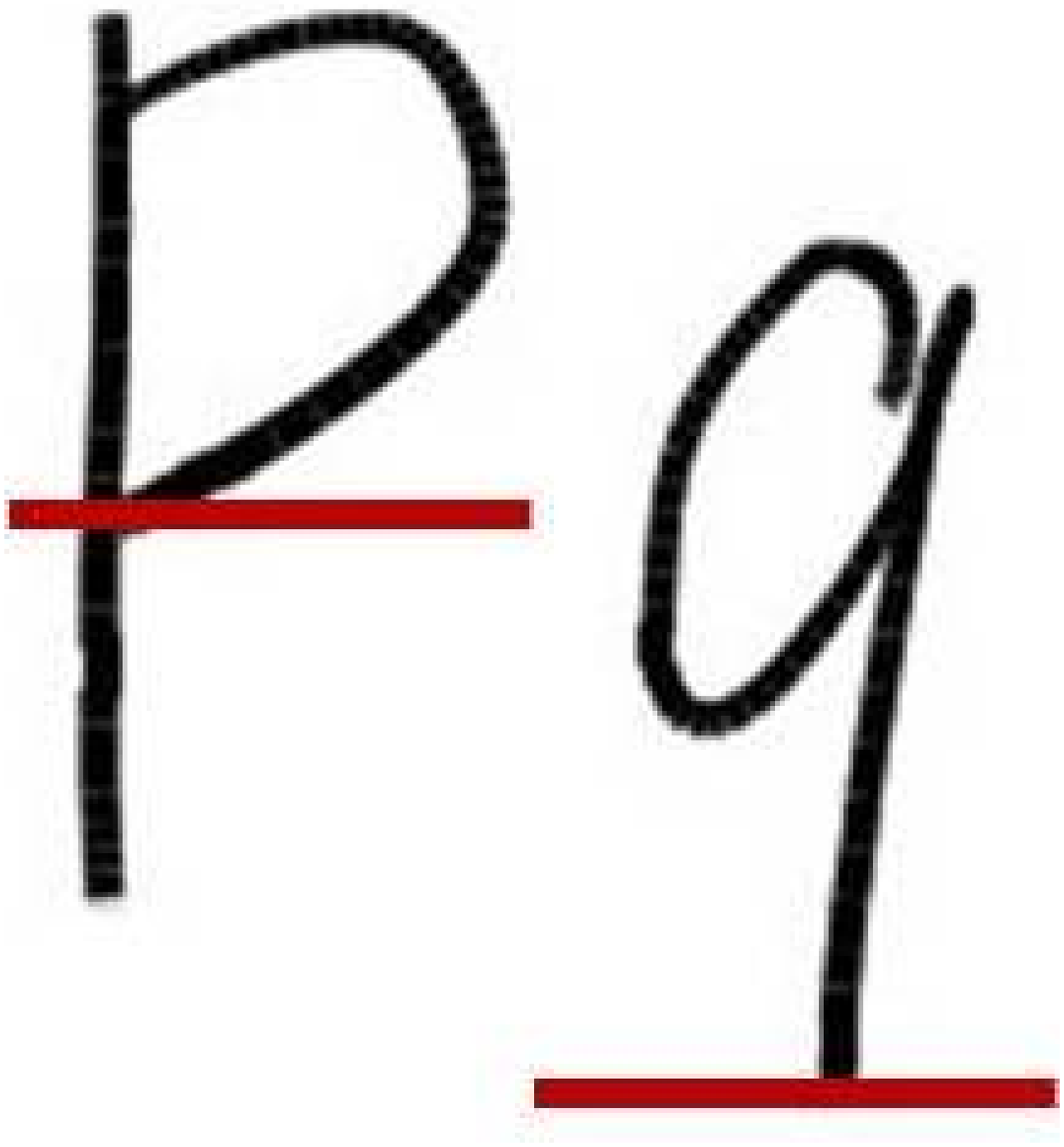}
\label{fig:p9-lowercase}
}
\caption[]{Disambiguation by baselines. (a) P9 (b) Pq (c) pq (d) p9}
\label{fig:juxtaposition-2}
\end{figure}

Juxtaposition ambiguity is common in mathematical handwriting recognition. This is usually
caused by symbols that are next to each other are written in different sizes and at different heights. 
Figure~\ref{fig:juxtaposition} shows an example with several relative positionings of two characters.  The first character can in each case be a ``P'' or ``p'' and the second can be interpreted as a ``q'' or ``9''. Together there could be a variety of possible interpretations:
$$
\begin{array}{ccccccc}
P^9 &  P9 &  P_9 &  ~~~ &  p^9 &  p9 &  p_9 \\
P^q &  Pq &  P_q &  ~~~ &  p^q &  pq &  p_q
\end{array}
$$
However, by comparing symbols' baseline locations and sizes, we can predict each expression with more confidence. 
This is because the baselines of subscripts and superscripts are typically placed 
slightly below or above the normal line of text and their sizes are relatively smaller.
Note that to determine the relative position, it is definitely not sufficient to compare the baselines of the symbol bounding boxes.  This is seen in Figure~\ref{fig:p9-lowercase}.  Similarly, having an imputed baseline determined by symbol class (such as at 50\% height for ``q'') is insufficient.
We thus find it is important to find and use the symbol's determining points.

\subsection*{Handwriting Neatening}
\begin{figure}[t]
\centering
\subfigure[]{
\includegraphics[width=4.4cm]{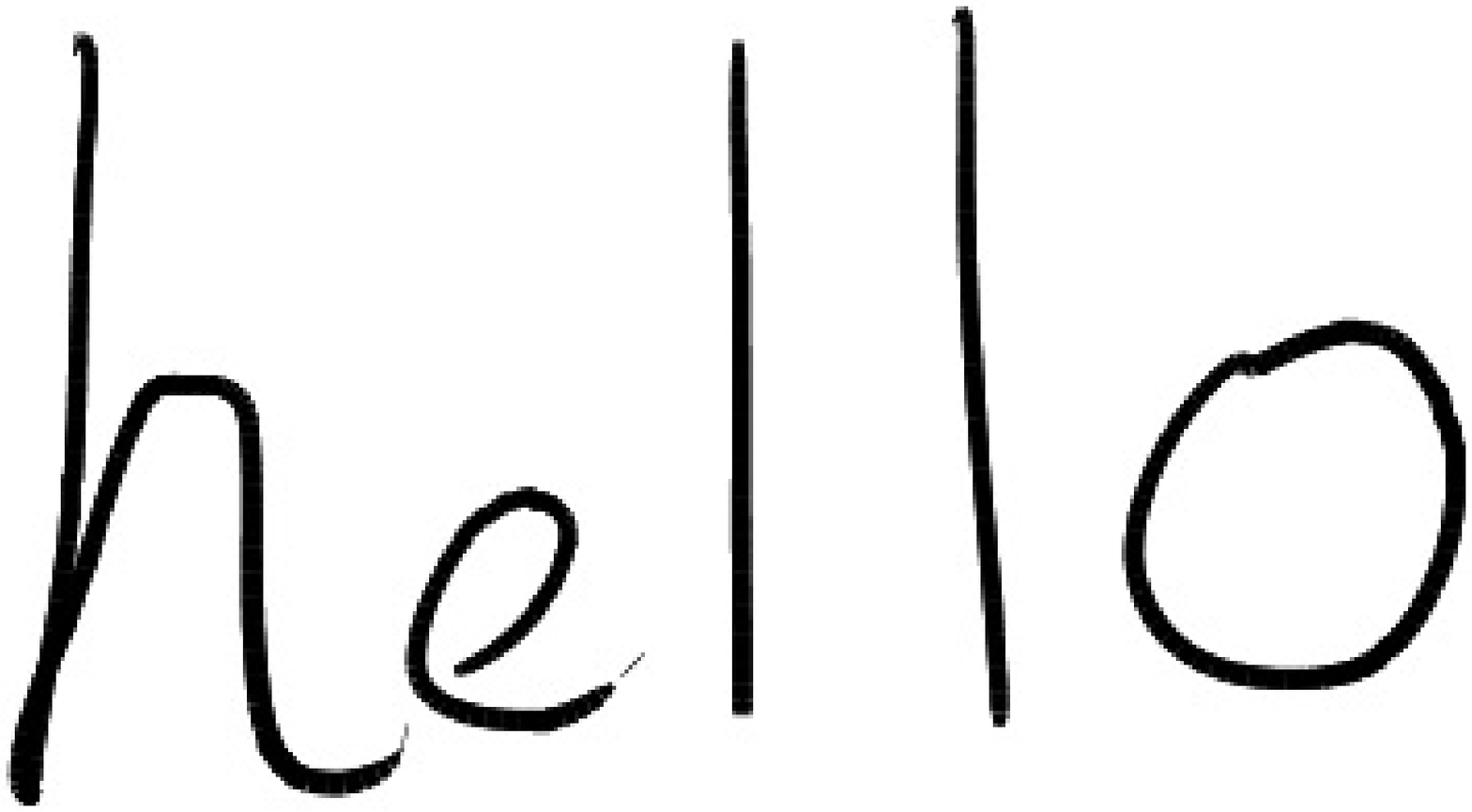}
\label{fig:hello-original}
}\hspace{15mm}
\subfigure[]{
\includegraphics[width=4.4cm]{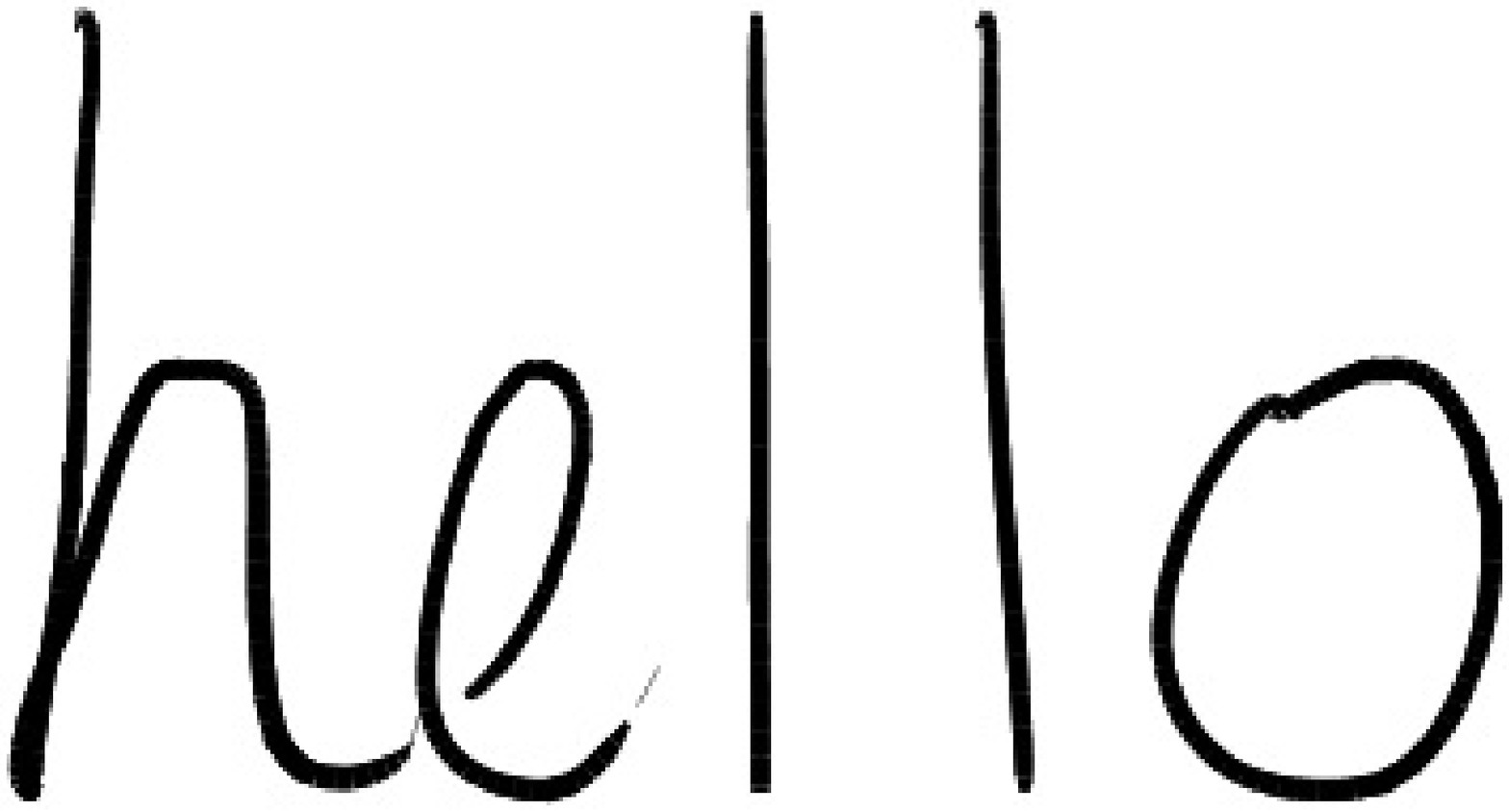}
\label{fig:hello-neatened}
}
\caption[]{Neatening  using determining points. (a) original, (b) neatened.}
\label{fig:use-case-neatening}
\end{figure}

\begin{figure}[t]
\centering
\vspace{\baselineskip}
\subfigure[]{
\includegraphics[width=5cm]{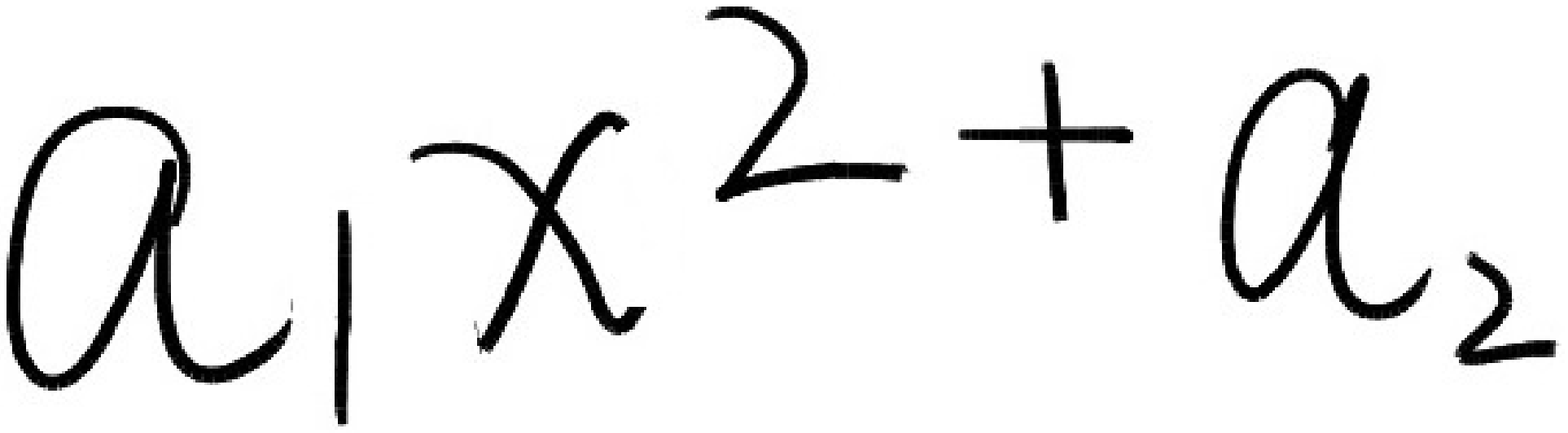}
\label{fig:math-original}
}\hspace{15mm}
\subfigure[]{
\includegraphics[width=5cm]{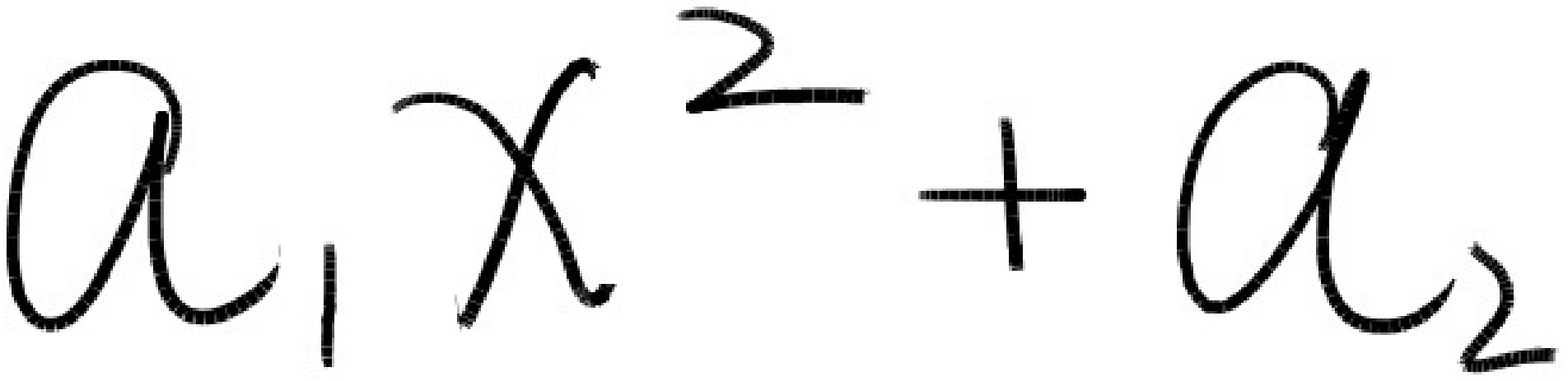}
\label{fig:/math-neatened}
}
\caption[]{Neatening  using determining points. (a) original, (b) neatened.}
\label{fig:use-case-neatening2}
\end{figure}

Handwriting neatening is becoming possible in some digital ink applications. It is used to transform handwriting to 
obtain visually appealing output while preserving the original writing style. Figure~\ref{fig:use-case-neatening}
shows an example. By identifying the determining points of each character, we can shift and scale these characters 
 to make corresponding metrics lines aligned properly, as shown in Figure~\ref{fig:hello-neatened}. 
Figure~\ref{fig:use-case-neatening2} shows a second example. In this case, all characters including the superscripts and subscripts 
were adjusted in order to obtain a normalized output. Transforming the function $y(s)$ for each symbol is the simplest approach to neatening.  A more aggressive approach is to replace each input symbol with the appropriately scaled version of the average of like symbols seen by the same writer, and further transformations can be employed.  However, this is beyond the scope of the present article.

\section{Conclusion and Future Work}\label{sec:conclusion}
We have presented an algorithm to identify automatically the determining 
points in handwritten symbols. Identifying
these determining points helps us better understand the scale
of individual characters as well as find the locations of 
certain desired features. In contrast to existing methods, which treat
digital ink traces as a collection of discrete points, this algorithm
relies on interpreting ink traces as single points in a functional
space. This allows device independence, on one hand, and a simple formulation of homotopic deformation, on the other. 

Various features can be recorded by using the determining point algorithm. The nature of the determining points depends on the symbol set used.
In our case, the symbols were based mostly on those of European languages and mathematical operators, so the baseline, x line, ascender line, descender line and cap line were used.

To evaluate the performance of the algorithm, we have tested it against a database of handwritten mathematical characters. The experiments showed promising
results. To demonstrate possible use of determining points, we have 
described two scenarios: handwriting recognition and handwriting
neatening, in both of which determining points have been found useful.

There are a few directions we would like to pursue in the
future. First, we wish to include determining points in our handwriting 
recognizer. It is expected that, combined with ambient baseline information, this will improve the recognition rate.
Secondly, we would like to investigate using rotation- and slant-invariant techniques~\cite{rotation-invariant,shear-invariant} in conjunction with the present methods.
At a more detailed-level, we would like to annotate all samples in our database using a supervised multi-step method.   This will allow us to perform a more satisfying statistical analysis of the effectiveness of our method.   Finally, before incorporating these techniques in our recognition framework, we would like to investigate the correlation between the model-sample distance and the number of steps required for low error rates, and how the number of required steps varies by class.

We would like to thank Isaac Watt for helping to organize the handwriting dataset
used in the experiments.

\newpage 
\nocite{}
\bibliography{references}   
\bibliographystyle{splncs}   

\end{document}